\documentclass[english]{article}
\pdfoutput=1
\usepackage[T1]{fontenc}
\usepackage[latin9]{inputenc}
\usepackage[a4paper]{geometry}
\usepackage{array}
\usepackage{wrapfig}
\usepackage{units}
\usepackage{multirow}
\usepackage{amsmath}
\usepackage{amssymb}
\usepackage{graphicx}
\usepackage{wasysym}

\makeatletter

\newcommand{\noun}[1]{\textsc{#1}}
\newcommand*\LyXZeroWidthSpace{\hspace{0pt}}
\providecommand{\tabularnewline}{\\}

\numberwithin{equation}{section}
\newcommand{\lyxaddress}[1]{
	\par {\raggedright #1
	\vspace{1.4em}
	\noindent\par}
}

\@ifundefined{date}{}{\date{}}
\makeatother

\usepackage{babel}
\begin{document}
\title{Limits to classification performance by relating Kullback-Leibler
divergence to Cohen's Kappa}
\author{L. Crow, S. J. Watts\thanks{Stephen.Watts@manchester.ac.uk}}
\maketitle

\lyxaddress{Department of Physics and Astronomy, The University of Manchester,
Oxford Road, Manchester, M13 9PL, UK}
\begin{abstract}
The performance of machine learning classification algorithms are
evaluated by estimating metrics, often from the confusion matrix,
using training data and cross-validation. However, these do not prove
that the best possible performance has been achieved. Fundamental
limits to error rates can be estimated using information distance
measures. To this end, the confusion matrix has been formulated to
comply with the Chernoff-Stein Lemma. This links the error rates to
the Kullback-Leibler divergences between the probability density functions
describing two classes. This leads to a key result that relates Cohen\textquoteright s
Kappa, $\kappa$, to the Resistor Average Distance, $R(P,Q)$, which
is the parallel resistor combination of the two Kullback-Leibler divergences.
The relation is, $\kappa=1-2^{-R(P,Q)}$. $R(P,Q)$ has units of bits
and is estimated from the same training data used by the classification
algorithm, using kNN estimates of the Kullback-Leibler divergences.
The classification algorithm gives the confusion matrix and $\kappa$.
Theory and methods are discussed in detail and then applied to Monte
Carlo data and real datasets. Four very different real datasets (Breast
Cancer, Coronary Heart Disease, Bankruptcy, and Particle Identification)
are analysed, with both continuous and discrete values, and their
performance compared to the expected theoretical limit given by the
formula. In all cases this analysis shows that the algorithms could
not have performed any better due to the underlying probability density
functions for the two classes. Important lessons are learnt on how
to predict the performance of algorithms for imbalanced data using
training datasets that are approximately balanced. Machine learning
is very powerful but classification performance ultimately depends
on the quality of the data and the relevance of the variables to the
problem.
\end{abstract}
\tableofcontents{}

\section{Introduction}

The performance of machine learning algorithms applied to classification
problems are evaluated by estimating parameters such as accuracy using
training data and cross-validation. These methods provide good estimates
of performance and are useful to compare the efficacy of different
algorithms. Most performance parameters can be extracted from the
confusion matrix which records how known classed instances are assigned
by the algorithm to a classification. A perfect algorithm would result
in a matrix with only diagonal entries. This subject is discussed
in detail in ref. \cite{Weka}. However, performance measures do not
prove that the best possible performance has been achieved for a given
dataset. They allow a comparison of different algorithms. This paper
will provide a methodology to establish whether a classification algorithm
has achieved the theoretical best case using information distance
measures. It will concentrate on the binary or two-class classification
problem. There is an extensive literature on how well one can classify
observations from $m$ possible classes dating back to the 1940's.
The Neyman-Pearson and Bayes Decision Rules are much discussed, e.g.
ref. \cite{StatPat}, with the Bayes Decision Rule for minimum error
leading to the concept of the Bayes Error Rate as the best one can
achieve. These ideas can be discussed in either a Bayesian or Frequentist
framework and lead to the same result. As Wasserman, ref. \cite{Allof}
, notes ``it is the nothing to do with Bayesian inference. The rule
can be obtained using either Frequentist or Bayesian methods.'' The
critical underlying idea is the likelihood ratio between the probability
density functions (pdfs) for the two classes. Consideration of this
ratio also led to the Kullback-Leibler Divergence, ref. \cite{Kullback},
Chernoff-Stein Lemma, ref. \cite{ChernoffStein} , and Chernoff Information,
ref. \cite{ChernoffInfo}, all results within an information-theoretic
framework, that are not dependent on prior probabilities. The relationship
between key information distance measures are shown in Fig. \ref{Figure1B}
which is an expanded version of the one first drawn in ref. \cite{Johnson}.
The figure includes the Renyi Divergence which will be shown in Section
3 to provide the key to solve the problem. If the pdf's are $p(x)$
and $q(x)$ for Class 1 and Class 2 instances, where a single variable
$x$ is used for clarity, but could also apply to a multivariate space,
then the information measures labelled in Fig. \ref{Figure1B} are
related to the \emph{exponential rates} for the optimal classification
performance, \cite{Johnson}. The Neyman-Pearson error for fixed $P_{M}$
is,
\begin{equation}
\lim_{N\rightarrow\infty}\frac{logP_{F}}{N}=-D(P\parallel Q)\label{Eq1:1}
\end{equation}
and the Chernoff Information, $C(P,Q)$, or bound on the Bayes Error
is
\begin{equation}
\lim_{N\rightarrow\infty}\frac{logP_{E}}{N}\leqq-C(P,Q)\label{Eq1:2}
\end{equation}
where, $P_{F}$ , $P_{M}$ and $P_{E}$ are the false alarm, miss
and average error probabilities respectively. Equ. \ref{Eq1:1} is
the Chernoff-Stein Lemma with $D(P\parallel Q)$ being the Kullback-Leibler
Divergence between the two pdf's. The Bhattacharyya Distance, $B(P,Q)$
ref. \cite{Bhattacharyya}, can also be used in Equ. \ref{Eq1:2}.
This bound is related to the Chernoff Information and $B(P,Q)\leqq C(P,Q)$.
For reference, in the Neyman-Pearson formulation, $P_{F}$ would be
a Type I Error or False Positive and $P_{M}$ would be a Type II Error
or False Negative. Since classification algorithms tend to treat classes
equally and search for a minimum error solution, this paper will not
use the Neyman-Pearson terminology. Equ. \ref{Eq1:1} and Equ. \ref{Eq1:2}
show that the probability of error depends exponentially on these
information divergence or distance measures. This will be used to
formulate the confusion matrix such that it follows the Chernoff-Stein
Lemma - Section 2.1. This matrix will be used to find a formula for
Cohen's Kappa, a key performance measure which is often used to judge
a classification algorithm - Section 2.2. One can then inter-relate
and link the parameters of the confusion matrix with the information
measures - Section 3.2. Finally, we estimate the information measures
from the training data - Section 3.3. One can then compare the algorithm
performance to the expected optimal prediction based on the information
measures, which depend only on the underlying pdfs. This comparison
delivers a verdict on whether the algorithm is performing as well
as the underlying pdfs will permit. The theory summarised in Section
3.4 is then applied successfully to four real datasets and two simulated
datsets in Section 4 and conclusions are given in Section 5. The methodology
just described in shown diagrammatically in Fig. \ref{Figure 2}.

\section{Theory - Confusion matrix and error rates}

First we will describe the confusion matrix for a two class problem
and ensure that it complies with the Chernoff-Stein Lemma. Then Cohen's
Kappa will be introduced and expressed in terms of exponential rate
parameters defining the confusion matrix. Note the units of the information
measures are bits and nats if the logarithm base, $b$, is $2$ or
$e$ respectively. The new formulation leads to important results
that relate the key exponential rates. 

\subsection{Confusion Matrix}

The confusion matrix will be written in a form that uses the Chernoff-Stein
Lemma, with exponential rates for the false-alarm probability. In
this section nats will be used. Consider a two class problem labelled
by 1 and 2. In the training set there are $N_{1}$ and $N_{2}$ entries
associated with these two classes. In a perfect world the classifier
would generate a confusion matrix, $\left[\begin{array}{cc}
N_{1} & 0\\
0 & N_{2}
\end{array}\right]$. However, in the real world, a classification algorithm would not
always correctly assign entries to the relevant class. The matrix
can be written as, $\left[\begin{array}{cc}
N_{1}^{T} & N_{1}^{L}\\
N_{2}^{L} & N_{2}^{T}
\end{array}\right]$ where, $N_{1}^{T}$and $N_{2}^{T}$ are the entries correctly (``True'')
assigned to Class 1 and 2 respectively, and $N_{1}^{L}$and $N_{2}^{L}$
are the entries that ``Leak'' into the wrong Class. Thus, $N_{1}^{T}+N_{1}^{L}=N_{1}$
and $N_{2}^{T}+N_{2}^{L}=N_{2}$. The total number of entries is,
$N=N_{1}+N_{2}$. This is shown diagramatically in Fig. \ref{Figure 3}.

A random decision algorithm would assign class on the basis of the
known number of entries in Class 1 and Class 2. The probability of
a Class 1 entry is, $\frac{N_{1}}{N_{1}+N_{2}}$. The probability
of a Class 2 entry is, $\frac{N_{2}}{N_{1}+N_{2}}$. Thus the ``random''
confusion matrix would be, $\left[\begin{array}{cc}
N_{1}\times\frac{N_{1}}{N} & N_{1}\times\frac{N_{2}}{N}\\
N_{2}\times\frac{N_{1}}{N} & N_{2}\times\frac{N_{2}}{N}
\end{array}\right]$. One now uses the Chernoff-Stein Lemma to describe the leaked entries
subject to the constraint that the matrix take the form required by
a random decision algorithm when the false-alarm rates are zero. The
matrix has to be of the form,

\begin{equation}
\left[\begin{array}{cc}
N_{1}^{T}=N_{1}-N_{1}^{L} & N_{1}^{L}=\frac{N_{1}N_{2}}{N}exp(-K_{12})\\
N_{2}^{L}=\frac{N_{1}N_{2}}{N}exp(-K_{21}) & N_{2}^{T}=N_{2}-N_{2}^{L}
\end{array}\right]\label{eq:3}
\end{equation}
$K_{12}$ and $K_{21}$ are the exponential rates that control the
``leakage'' of entries from Class 1 to Class 2 and from Class 2
to Class 1 respectively. They will be shown to be related to the Kullback-Leibler
Divergence for the probability density functions describing the Class
1 and Class 2 entries.

\subsection{Cohen's Kappa}

To date, over 26 performance measures, ref. \cite{RClass}, have been
derived from the $2\times2$ confusion matrix, although it only has
two free parameters. We apply the matrix, Equ. \ref{eq:3}, to calculate
Cohen's Kappa, ref. \cite{Cohen}, which was originally intended to
assess agreement between two judgements in psychological measurement.
To quote ref. \cite{Cohen}, it is the ``proportion of agreement
after chance agreement is removed''. When used in binary classification,
it gives the efficiency of the algorithm after correcting for chance
agreement. The confusion matrix defined in Equ. 2.1 is carefully formulated
to take random agreement into account. The expression for Cohen's
Kappa, $\kappa$, is
\begin{equation}
\kappa=\frac{p_{o}-p_{e}}{1-p_{e}}\label{eq:2}
\end{equation}
where, $p_{o}$ is the observed agreement and $p_{e}$ is the agreement
that one would find by random expectation. Note that $1-\kappa$ is
the average probability that the classification is in error. $\kappa$
can now be written in terms of the confusion matrix, Equ. \ref{eq:3}.
The observed agreement is, $p_{0}=\frac{N_{1}^{T}+N_{2}^{T}}{N}$.
Random expectation for Class 1 is, $p_{1}=\frac{N_{1}}{N}\times\frac{N_{1}^{T}+N_{2}^{L}}{N}$.
Random expectation for Class 2 is, $p_{2}=\frac{N_{2}}{N}\times\frac{N_{2}^{T}+N_{1}^{L}}{N}$.
Then $p_{e}=p_{1}+p_{2}$. Using these relations and after some algebra,
one arrives at the relationship between $\kappa$ and the parameters
used in the matrix, Equ. \ref{eq:3}.

\begin{equation}
\kappa=\frac{2N_{1}(1-exp(-K_{21}))+2N_{2}(1-exp(-K_{12}))}{N_{1}(2+exp(-K_{12})-exp(-K_{21}))+N_{2}(2+exp(-K_{21})-exp(-K_{12}))}\label{eq:3-1}
\end{equation}

There are two important limits for Equ. \ref{eq:3-1}
\begin{enumerate}
\item When the classes are balanced, $N_{1}=N_{2}=\frac{N}{2}$ , $\kappa=1-\frac{1}{2}(exp(-K_{12})+exp(-K_{21}))$.
This only depends on two parameters and makes $\kappa$ a robust statistic
because there is no dependence on $N_{1}$ or $N_{2}$.
\item When $K_{12}=K_{21}$$\equiv K$ ,
\begin{equation}
\kappa=1-exp(-K)\label{eq:2.4}
\end{equation}
independent of $N_{1}$ and $N_{2}$, e.g. imbalanced classes. Again
$\kappa$ is a robust statistic. This situation is not uncommon because
a classification algorithm will do its best to optimize the efficiency
for all classes, which will tend to make the rates similar. 
\end{enumerate}
The $\kappa$ statistic has been criticised because of its dependence
on how the $N$ entries are balanced between $N_{1}$ and $N_{2}$.
Later it will be shown that this is inevitable and that this parameter
correctly describes the classification performance, especially when
the algorithm is looking to correctly classify both classes. Kappa
is a key statistic describing the overall performance of the classification
algorithm. Equ. 2.4 can be used to define the overall exponential
error rate in terms of one parameter, $K$ defined as,

\begin{equation}
K\equiv-log_{b}(1-\kappa)\label{eq:2.5}
\end{equation}

\subsection{Inter-relations between leakage rates and overall error rate}

Once a classification algorithm has been applied to a training dataset,
the resultant confusion matrix can be used to derive the three key
parameters. The actual $N_{1}^{L}$ and $N_{2}^{L}$ elements can
be used to estimate the $K_{12}$ and $K_{21}$ exponential leakage
rates respectively using the formulae below and derived from Equ.
\ref{eq:3}.

\begin{equation}
K_{12}=-log_{b}\left(\frac{1}{f_{1}f_{2}N}\cdot N_{1}^{L}\right)\label{eq:2.6}
\end{equation}

\begin{equation}
K_{21}=-log_{b}\left(\frac{1}{f_{1}f_{2}N}\cdot N_{2}^{L}\right)\label{eq:2.7}
\end{equation}
where, $f_{1}\equiv\frac{N_{1}}{N}$ and , $f_{2}\equiv\frac{N_{2}}{N}$
are the fraction of entries in Class 1 and Class 2 respectively. For
this two class problem, $f_{1}+f_{2}=1$. The overall rate, $K$ is
given by Equ. (2.5) and is estimated from a calculation of Cohen's
Kappa. Since the confusion matrix has four elements and is constrained
by knowledge of the number of Class 1 and Class 2 entries, there can
only be two free parameters, and these three rates are related. To
understand this, re-write Equ. \ref{eq:3-1} using Equ. \ref{eq:2.5}
on the left-hand side,
\begin{equation}
1-exp(-K)=\frac{2N_{1}(1-exp(-K_{21}))+2N_{2}(1-exp(-K_{12}))}{N_{1}(2+exp(-K_{12})-exp(-K_{21}))+N_{2}(2+exp(-K_{21})-exp(-K_{12}))}\label{eq:2.8}
\end{equation}
Simplify this by assuming that $\left|K_{12}-K\right|$ and $\left|K_{21}-K\right|$$\apprle0.4$
nats or $0.6$ bits. Then multiply the numerator and denominator of
Equ. \ref{eq:2.8} by $exp(K)$. After applying a Taylor expansion
approximation to the $exp(K_{12}-K)$ and $exp(K_{21}-K)$ terms and
some algebra one finds, 
\begin{equation}
K\backsimeq f_{1}K_{12}+f_{2}K_{21}+\frac{1}{2}(1-exp(-K))(K_{21}-K_{12})(f_{1}-f_{2})\label{eq:2.9-1}
\end{equation}
There are two solutions to Equ. \ref{eq:2.9-1}
\begin{enumerate}
\item For balanced classes, $f_{1}=f_{2}=\frac{1}{2}$ , $K\backsimeq f_{1}K_{21}+f_{2}K_{12}$=$\frac{1}{2}(K_{12}+K_{21})$. 
\item For symmetric Kullback-Leibler, $K_{12}=K_{21}$, $K\simeq f_{1}K_{21}+f_{2}K_{12}$.
In general, the last term is small compared to the first two factors,
so one arrives at the same result.
\end{enumerate}
In conclusion, there is an approximate relation between the three
rates. This is exact for balanced classes or symmetric Kullback-Leibler
divergence. 

\begin{equation}
K_{W}\equiv f_{1}K_{21}+f_{2}K_{12}\simeq K\label{eq:2.9}
\end{equation}
$K_{W}$ will be referred to as the \emph{weighted error rate}. The
confusion matrix is described in terms of these three parameters.
These are the exponential average error rate $(K)$, exponential leakage
rate causing Class 1 entries to be misclassified as Class 2 $(K_{12})$,
exponential leakage rate causing Class 2 entries to be misclassified
as Class 1 $(K_{21})$. As these descriptions are rather long, they
will be referred to as the\emph{ average error rate, $K$ , and leakage
rates $K_{12}$ and $K_{21}$. }Equ. \ref{eq:2.9} is a very important
result. It is a rate equation because it relates these three rates
and drives the classification algorithm which is designed to give
the best value of $K$ which itself is constrained by the Chernoff
bound. 

\LyXZeroWidthSpace{}

One final result from this formulation of the confusion matrix is
that there is a limit to the leakage rate estimate. The smallest value
of the off-diagonal entries in Equ. \ref{eq:2.6} or Equ. \ref{eq:2.7}
is one. This puts a limit on the maximum leakage rate,
\begin{equation}
K_{Max}^{L}=log_{b}N+log_{b}(f_{1}f_{2})\label{eq:2.10}
\end{equation}
This confirms a result in ref. \cite{logN}, which found a $\mathcal{O}log_{b}(N)$
limit on the Kullback-Leibler divergence. As this section shows, this
is a consequence of the Chernoff-Stein Lemma.

\section{Theory - information divergence measures and error rates}

This section will derive in detail the relationships between the error
rates and the information divergence measures and then relate these
to the parameters, $K$, $K_{12}$, and $K_{21}$ of the confusion
matrix. Next, it will show how to estimate the information distance
measures independently from training data. The final sub-section returns
to the overall methodology of the paper, which is shown schematically
in Fig. \ref{Figure 2}, linking and referencing the key equations.

\subsection{The Renyi Divergence, Resistor Average Distance, and the overall
average error rate}

\begin{figure}[ph]
\includegraphics[scale=0.8]{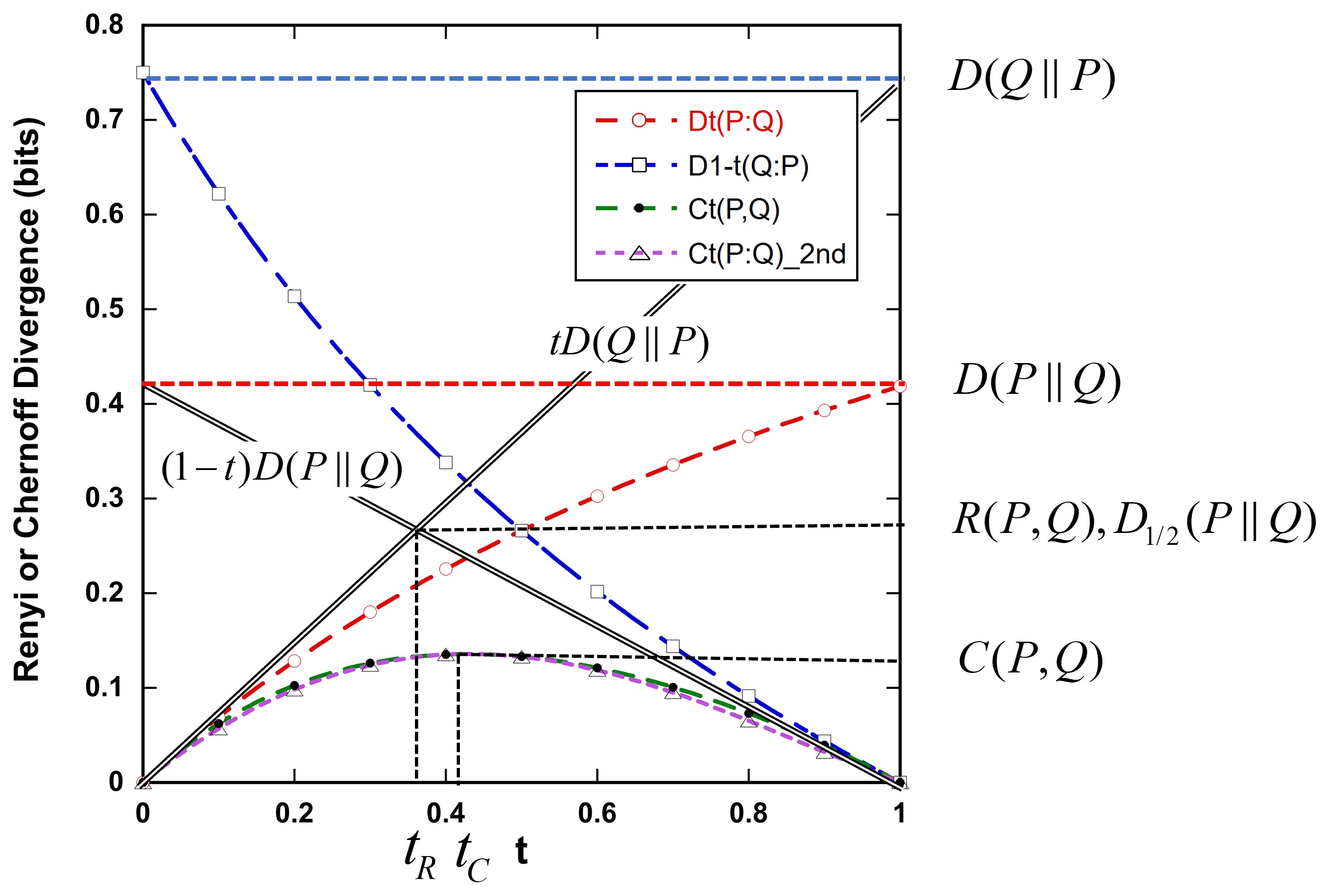}

\caption{This figure is an update of one used in ref. \cite{Johnson} to show
the relationships between key information theoretic distances. The
curves are for a 1D exponential model described in Section 4.1. Renyi
Divergences, $D_{t}(P\parallel Q)$ and $D_{1-t}(Q\parallel P)$,
Chernoff Divergence, $C_{t}(P\parallel Q)$, Kullback-Leibler Divergences,
$D(P\parallel Q)$ and $D(Q\parallel P)$. The Resistor Average Distance,
$R(P,Q)$, \cite{Johnson}, is at $t=t_{R}$ and Chernoff Information,
$C(P,Q)$, is at $t=t_{C}$. $R(P,Q)$ is the value at which the two
double lines meet. These are tangential to the Chernoff Divergence
at $t=0$ and $t=1$. The Renyi Divergence at $t=1/2$, $D_{1/2}(P\parallel Q)=D_{1/2}(Q\parallel P)$,
has a value close to $R(P,Q).$ This is not an accident as the main
text explains. Not shown on figure, but for reference, the Bhattacharyya
Distance, $B(P,Q)$ is the value of the Chernoff Divergence at $t=$$1/2$
, which is $\frac{1}{2}D_{1/2}(P\parallel Q)$. A second order approximation
to the Renyi divergence is used to estimate the Chernoff Divergence,
which is shown in the figure. See Section 3.1 for details.}

\label{Figure1B}
\end{figure}

Equ. \ref{Eq1:1} relates the leakage rate to the KullBack-Leibler
divergence, which is a consquence of the Chernoff-Stein Lemma. The
assumption of exponential error rates has been used in the formulation
of the confusion matrix. However, one can only apply the Kullback-Leibler
Divergence to a situation in which the classifier wishes to get the
best performance for one class to the detriment of the other. This
situation applies in signal detection when there is no interest in
the noise. This corresponds to $t=0$ and $t=1$ in Fig. \ref{Figure1B}.
Chernoff in ref. \cite{ChernoffInfo} realized that when treating
each class equally, the answer lay between $0\leqq t\leqq1$ with
the Chernoff Information as a bound on the overall error rate. Ref.
\cite{Johnson} proposed the Resistor Average Distance, $R(P,Q)$,
as an estimate for this bound. This is the parallel resistor combination
of $D(P\parallel Q)$ and $D(Q\parallel P).$ It corresponds to the
point where the two straight lines in Fig. \ref{Figure1B} meet. These
lines are the tangents to the Chernoff Divergence at $t=0$ and $t=1$
which have a slope of $D(Q\parallel P)$ and $D(P\parallel Q)$ respectively.
In this section, the motivation for $R(P,Q)$ will be explained in
terms of the Renyi Divergence rather than using the argument in Ref.
\cite{Johnson}. The Renyi Divergence is a generalization of the Kullback-Leibler
Divergence and is described in detail in ref. \cite{vanErven}, which
also describes the hypothesis testing problem in terms of the Renyi
Divergence. It is defined as follows,

\begin{equation}
D_{t}(P\parallel Q)\equiv\frac{1}{t-1}log\int_{S}p(x)\left(\frac{p(x)}{q(x)}\right)^{t-1}dx=\frac{1}{t-1}log\int_{S}p(x)^{t}q(x)^{1-t}dx
\end{equation}
The middle form clearly shows the link to the likelihood ratio, $\frac{p(x)}{q(x)}$.
The final form is the one most often used. The log term is due to
Chernoff from which the overall error bound was derived. Nowadays,
it is often referred to as the Chernoff Divergence, ref. \cite{Crooks},
which is,

\begin{equation}
C_{t}(P\parallel Q)=(t-1)D_{t}(P\parallel Q)
\end{equation}
The Chernoff Information or Distance is,
\begin{equation}
C(P,Q)\equiv Max(C_{t}(P\parallel Q))
\end{equation}
This occurs at a value of $t=t_{c}$, which is indicated in Fig. \ref{Figure1B}.
Only values of the Renyi and Chernoff Divergence for $0\leqq t\leqq1$
are of interest in this paper, in which case, $C_{t}(P\parallel Q)=C_{1-t}(Q\parallel P)$.
See ref. \cite{vanErven} for other values of $t.$ From the definitions
it is easy to show that,
\begin{equation}
(1-t)D_{t}(P\parallel Q)=tD_{1-t}(Q\parallel P)=C_{t}(P\parallel Q)\label{eq:3.4}
\end{equation}
This is called Skew Symmetry, ref. \cite{vanErven}, and is a key
property of the Renyi Divergence from which the concept of Resistor
Average Distance appears naturally. Finally, the Kullback-Leibler
Divergence is,
\begin{equation}
D(P\parallel Q)\equiv\int_{S}p(x)log\left(\frac{p(x)}{q(x)}\right)dx
\end{equation}
which is $D_{t}(P\parallel Q)$ at $t=1$. Note also that $D(Q\parallel P)=D_{1-t}(Q\parallel P)$
at $t=0$. The relationships between Kullback-Leibler Divergence and
Chernoff Divergence are shown in Fig. \ref{Figure1B}. The Resistor
Average Distance is defined mathematically as,
\begin{equation}
(1-t)D(P\parallel Q)=tD(Q\parallel P)=R(P,Q)\label{eq:3.6}
\end{equation}
Equ. \ref{eq:3.6} is of the same form as the Skew Symmetry property
given in Equ. \ref{eq:3.4}. These have similar solutions. The Resistor
Average Distance is the point, $t=t_{R}$, at which the two tangential
lines meet in Fig. \ref{Figure1B},

\begin{equation}
R(P,Q)=\frac{D(P\parallel Q)D(Q\parallel P)}{D(P\parallel Q)+D(Q\parallel P)}
\end{equation}
 or 
\begin{equation}
\frac{1}{R(P,Q)}=\frac{1}{D(P\parallel Q)}+\frac{1}{D(Q\parallel P)}\label{eq:3.8}
\end{equation}
The solution for the Chernoff Divergence for $0\leqq t\leqq1$,

\begin{equation}
C_{t}(P\parallel Q)=\frac{D_{t}(P\parallel Q)D_{1-t}(Q\parallel P)}{D_{t}(P\parallel Q)+D_{1-t}(Q\parallel P)}\label{eq:3.9}
\end{equation}
or
\begin{equation}
\frac{1}{C_{t}(P,Q)}=\frac{1}{D_{t}(P\parallel Q)}+\frac{1}{D_{1-t}(Q\parallel P)}
\end{equation}
To link $R(P,Q)$ with the Chernoff Divergence, take a first-order
approximation to the Renyi Divergence that $D_{t}(P\parallel Q)\backsimeq tD(P\parallel Q)$
and $D_{1-t}(Q\parallel P)\simeq(1-t)D(Q\parallel P).$ This is an
equality for a symmetric Kullback-Leibler Divergence, i.e $D(P\parallel Q)=D(Q\parallel P)$.
This approximates the Renyi Divergences with two lines which are related
to the tangential lines in Fig. \ref{Figure1B} when exchanging $t\longleftrightarrow1-t$
and $P\leftrightarrow Q$. Substituting in Equ. \ref{eq:3.9} gives,

\begin{equation}
C_{t}(P\parallel Q)=\frac{tD(P\parallel Q)(1-t)D(Q\parallel P)}{tD(P\parallel Q)+(1-t)D(Q\parallel P)}\label{eq:3.11}
\end{equation}
Evaluate Equ. \ref{eq:3.11} at $t=0.5$ which is the Bhattacharyya
Distance, which is known to be close the Chernoff Distance, then,

\begin{equation}
C_{1/2}(P\parallel Q)=B(P,Q)=\frac{R(P,Q)}{2}\label{eq:3.12}
\end{equation}
For a symmetric Kullback-Liebler Divergence, $D(P\parallel Q)=D(Q\parallel P)$

\begin{equation}
R(P,Q)=2C_{1/2}(P,Q)=2C(P,Q)=2B(P,Q)=D_{1/2}(P\parallel Q)=\frac{D(P\parallel Q)}{2}\label{eq:3.13}
\end{equation}
A second order fit to the Renyi Divergence can be written as,
\begin{equation}
D_{t}(P\parallel Q)=tD(P\parallel Q)+At(1-t),D_{1-t}(Q\parallel P)=(1-t)D(Q\parallel P)+Bt(1-t)\label{eq:3.14-1}
\end{equation}
This is the only form compatible with the boundary conditions at $t=0$
and $t=1$. Since these divergences meet at $t=0.5$ one finds that,
\begin{equation}
A=4D_{1/2}(P\parallel Q)-2D(P\parallel Q),B=4D_{1/2}(Q\parallel P)-2D(Q\parallel P)\label{eq:3.15}
\end{equation}
The curves drawn in Fig. \ref{Figure1B} are for the exponential model
described in Section 4.1 for which $A=0.2269$ and $B=-0.4342$. Simple
algebra starting with Equ. \ref{eq:3.15} finds that,
\begin{equation}
D_{1/2}(P\parallel Q)=D_{1/2}(Q\parallel P)=R(P,Q)\left[1+\frac{A}{4D(P\parallel Q)}+\frac{B}{4D(Q\parallel P)}\right]\label{eq:3.16-1}
\end{equation}
Since $A$ and $B$ are small compared to $D(P\parallel Q)$ and $D(Q\parallel P)$
respectively, and have opposite sign, $R(P,Q)$ is a good estimate
of the Renyi Divergence at $t=0.5.$ The correction term in square
brackets in Equ. \ref{eq:3.16-1} for the exponential model is $0.991$.
The second order approximation to the Renyi divergence, Equ. \ref{eq:3.14-1}
is used to estimate the Chernoff divergence, Equ. \ref{eq:3.11},
and this is shown in Fig. \ref{Figure1B} . There is good agreement
with the calculated Chernoff divergence. Equ. \ref{eq:3.16-1} also
suggests that $D_{1/2}(P\parallel Q)$ is the information distance
measure relevant to the overall error rate since it is half way between
the solutions at $t=0$ ($D(Q\parallel P)$ ) and $t=1$ ($D(P\parallel Q)$)
which correspond to attaining the best error for just one of the classes.
In mathematical terms, the \emph{conjecture} is that,

\begin{equation}
\lim_{N\rightarrow\infty}\frac{logP_{E}}{N}\simeq-D_{1/2}(P\parallel Q)\simeq-R(P,Q)\label{eq:3.17}
\end{equation}
One can also write Equ. \ref{Eq1:1} as,
\begin{equation}
\lim_{N\rightarrow\infty}\frac{logP_{F}}{N}=-D_{1}(P\parallel Q)=D(P\parallel Q)
\end{equation}
The value of $t$ at which $R(P,Q)$ occurs, Fig. \ref{Figure1B},
is

\begin{equation}
t_{R}=\frac{D(P\parallel Q)}{D(P\parallel Q)+D(Q\parallel P)}\label{eq:3.19}
\end{equation}
This conjecture is borne out by application to data in Section 4 and
a mathematical result in the next section which shows from the confusion
matrix parameterisation, underpinned by the Chernoff-Stein Lemma,
that $K$ at $t=t_{R}$ is $R(P,Q)$. Referring back to ref. \cite{ChernoffInfo}
and using modern terminology, it notes that the Chernoff Divergence
and Kullback-Leibler Divergence are two different functionals on a
curve relating Type 1 and Type 2 errors for likelihood ratio tests
. The Renyi Divergence is another functional which provides the link
between the Type 1 and Type 2 errors and also the Bayes Error when
the classifier works well on both classes. 

\subsection{Link between the confusion matrix parameters and information theoretic
distances}

Associate $P$ and $Q$ with Class 1 and Class 2 respectively. The
two rates that define the confusion matrix are $K_{12}$ and $K_{21}$,
so the classifier will minimise the overall average error rate by
keeping $K\simeq R(P,Q)$ but must ensure that $K\simeq f_{1}K_{21}+f_{2}K_{12}$
which was shown earlier, Equ. \ref{eq:2.9} . We will consider two
cases.
\begin{enumerate}
\item The Kullback-Leibler Divergence is symmetric. The Renyi Divergence
will be linear and Equ. \ref{eq:3.13} will apply. Thus $K=D_{1/2}(P\parallel Q)=R(P,Q)=\frac{D(P\parallel Q)}{2}=\frac{D(Q\parallel P)}{2}$.
This occurs at $t=0.5$ and if the classes are balanced then, $K_{12}=K_{21}=K$
due to Equ. \ref{eq:2.9}. This is an exact solution. If the classes
are imbalanced, $f_{1}\neq f_{2}$ then $K_{12}\neq K_{21}$ but the
classification algorithm will try to preserve the value of $K$. This
can be understood by considering the extremes. When extrapolating
$f_{1}\rightarrow1$ there are only Class 1 entries. In this case,
one applies the Chernoff-Stein lemma, Equ. \ref{Eq1:1}, and $K_{12}\simeq D(Q\parallel P)$,
which corresponds to $t=1.$ When extrapolating, $f_{1}\rightarrow0$
there are only Class 2 entries, and in this case, one applies the
Chernoff-Stein lemma, Equ. \ref{Eq1:1}, and $K_{21}\simeq D(P\parallel Q)$,
which corresponds to $t=0$. In this case, $D(P\parallel Q)=D(Q\parallel P)$,
but the argument applies to the asymmetric case too. One is thus led
to write the following formulae for $K_{12}$ and $K_{21}$,
\begin{equation}
K_{12}\sim D(2,1)f_{1},K_{21}\sim D(1,2)f_{2}\label{eq:3.16-2}
\end{equation}
This provides a solution to the classification problem for the three
key values of $t=0$, $t=0.5$, and $t=1$, which correspond respectively
to the Chernoff-Stein, Chernoff Bound, and Chernoff-Stein solutions
and $t$ directly maps to the fraction of entries in Class 1, $f_{1}$.
The parameters $D(1,2)$ and $D(2,1)$ are estimated from data and
should correspond to the underlying values $D(P\parallel Q)$ and
$D(Q\parallel P)$ respectively. Fig. \ref{Figure1B} is the key to
this conclusion. Combining Equ. \ref{eq:2.9} and Equ. \ref{eq:3.16-2}
one finds that,
\begin{equation}
K\simeq K_{W}=f_{1}f_{2}D(1,2)+f_{2}f_{1}D(2,1)=f_{1}(1-f_{1})(D(1,2)+D(2,1))\label{eq:3.21}
\end{equation}
This immediately shows why one can preserve a constant $K$ and that
the maximum value of $K$ will be at $f_{1}=$$\frac{1}{2}$. There
will be a region of $f_{1}$ when $K$ will be \emph{roughly }constant
and the classification will apply to both classes. At the extremes,
when the classes are severely imbalanced, this may no longer apply.
This will be discussed later. Results from simulations, shown later,
indicate that one needs to modify Equ. \ref{eq:3.16-2}, with an extra
correction term, that is linked to the overlap between the two classes.
\begin{equation}
K_{12}=\Delta_{1}+(D(2,1)-\Delta_{1})f_{1}\label{eq:3.17-1}
\end{equation}
\begin{equation}
K_{21}=\Delta_{2}+(D(1,2)-\Delta_{2})f_{2}=D(1,2)-(D(1,2)-\Delta_{2})f_{1}\label{eq:3.18-1}
\end{equation}
 where, $0\leq\Delta_{1}\leq D(2,1)$ and $0\leq\Delta_{2}\leq D(1,2)$.
For the symmetric case, $D(1,2)=D(2,1)$, $\Delta_{1}=\Delta_{2}$
and $K_{12}=K_{21}$. The value of $f_{1}$ when $K_{12}=K_{21}$
will be called the \emph{balance point}, which in this case is, $f_{B}=0.5$.
Despite the more complicated equations for $K_{12}$ and $K_{21}$,
it is simple to show that the maximum value of $K$ is still at $f_{1}=0.5$
provided $\Delta_{1}$ and $\Delta_{2}$ are much smaller than $D(1,2)$
and $D(2,1)$. 
\item The Kullback-Leibler Divergence is not symmetric. The overall error
rate will still be controlled by $K=R(P,Q).$ Equ. \ref{eq:3.17-1}
and Equ. \ref{eq:3.18-1} will still apply. This balance point can
be approximated well as,
\begin{equation}
f_{B}\simeq\frac{D(1,2)}{D(1,2)+D(2,1)}\left[1+\frac{\Delta_{1}+\Delta_{2}}{D(1,2)+D(2,1)}-\frac{\Delta_{1}}{D(1,2)}\right]\label{eq:3.24}
\end{equation}
This works well because $\Delta_{1}$ and $\Delta_{2}$ are small
compared to $D(1,2)$ and $D(2,1)$. Moreover, one can see immediately
that,
\begin{equation}
f_{B}\approx t_{R}
\end{equation}
The balance point for the two error rates is the same as the value
of $t$ at which the Resistor Average Distance is determined. This
further justifies the link between the $f_{1}$ and $t.$ One can
estimate the value of $K$ at the balance point. To clearly illustrate
the result, the problem will be simplified with $\Delta_{1}=\Delta_{2}=0.$
At the balance point, $f_{1}=\frac{D(1,2)}{D(1,2)+D(2,1)}$ and $f_{2}=\frac{D(2,1)}{D(1,2)+D(2,1)}$.
Putting these into Equ. \ref{eq:3.21}, one finds,
\begin{equation}
K\simeq K_{W}=\frac{D(1,2)}{D(1,2)+D(2,1)}\bullet\frac{D(2,1)}{D(1,2)+D(2,1)}\bullet(D(1,2)+D(2,1))=\frac{D(1,2)D(2,1)}{D(1,2)+D(2,1)}=R(P,Q)\label{eq:3.26}
\end{equation}
This returns us to the start of this section, and justifies the underlying
model that the algorithm will try its best to maintain $K=R(P,Q)$
while keeping $K\simeq f_{1}K_{21}+f_{2}K_{12}$. This is an approximation
but works well in the region of $f_{1}$ when it is possible to classify
both classes adequately. This region will depend upon the underlying
pdfs. The application section will show how one can determine the
region in $f_{1}$ that permits a reasonable two-class classification.
Using units of bits, an equation that can be checked for all data,
re-writes Equ. \ref{eq:2.4} as,
\begin{equation}
\kappa\simeq1-2^{-R(P,Q)}=1-2^{-CDR}\label{eq:3.27}
\end{equation}
This important equation connects the classification performance with
that expected from the independently estimated information distance
measure for $R(P,Q).$ $CDR$ is the $k^{th}$ nearest-neighbour (kNN)
estimate of $R(P,Q)$ which is described in Section 3.3. It should
work well at the balance point and more generally in the region where
the algorithm can classify both classes. 
\end{enumerate}

\subsection{Estimating the Kullback-Leibler Divergence from training data}

The Kullback-Leibler Divergence is estimated from the same training
data used by the classification algorithm but with an independent
method that only considers the underlying pdfs. The use of kNN non-parametric
estimators of entropy was initiated by the work of ref. \cite{Kozachenko}.
The application to divergence was made by ref. \cite{Wang}. These
methods produce an estimate that is consistent and unbiased. Assigning
Class 1 and Class 2 entries to distributions $P$ and $Q$ respectively
and using ref. \cite{Wang} we write the estimate of $D(P\parallel Q)$
in bits as,
\begin{equation}
CDI(1,2)=\frac{d}{N_{1}}\mathop{\sum_{i=1}^{i=N_{1}}log_{2}\left(\frac{\lambda_{i}^{12}}{\lambda_{i}^{1}}\right)}+log_{2}\left(\frac{N_{2}}{N_{1}-1}\right)\label{eq:3.16}
\end{equation}
This is called the Class Distance Indicator between Class 1 and Class
2 entries to make it clear that it is an \emph{estimate} of the Kullback-Leibler
Divergence beween these two distributions. Its uses the kNN distances
which are defined as follows,
\begin{itemize}
\item $\lambda_{i}^{1}$ is the nearest neighbour distace between the $i^{th}$
point in Class 1 and all the other Class 1 points.
\item $\lambda_{i}^{12}$ is the nearest neighbour distance between the
$i^{th}$ point in Class 1 and all Class 2 points.
\item $d$ is the number of variables or dimensionality of the data space.
\end{itemize}
Equ. \ref{eq:3.16} deals automatically with imbalanced classes due
to the second term. This equation is applied to binned data with the
bin value converted to a real value by adding a uniform random number
between zero and one. This allows one to use the same algorithm for
discrete, continuous and mixed data. Moreover, the $CDI$ can be estimated
multiple times and then averaged. This reduces the error and provides
a lower bound on the error for one computation. For continuous data
the binning uses an algorithm that ensures each variable has the same
Shannon entropy. This reduces systematic errors. A detailed paper
is in preparation but one can obtain further information on the method
and its performance in refs. \cite{Crow,WattsCrow,Hist}. The Resistor
Average Distance, $R(P,Q)$, is estimated by a parameter called, $CDR$,
which is the parallel resistor combination of $CDI(1,2)$ and $CDI(2,1),$
\begin{equation}
\frac{1}{CDR}=\frac{1}{CDI(1,2)}+\frac{1}{CDI(2,1)}\label{eq:3.20}
\end{equation}

\subsection{Summary of methodology and key equations}

\begin{figure}
\includegraphics[scale=0.5]{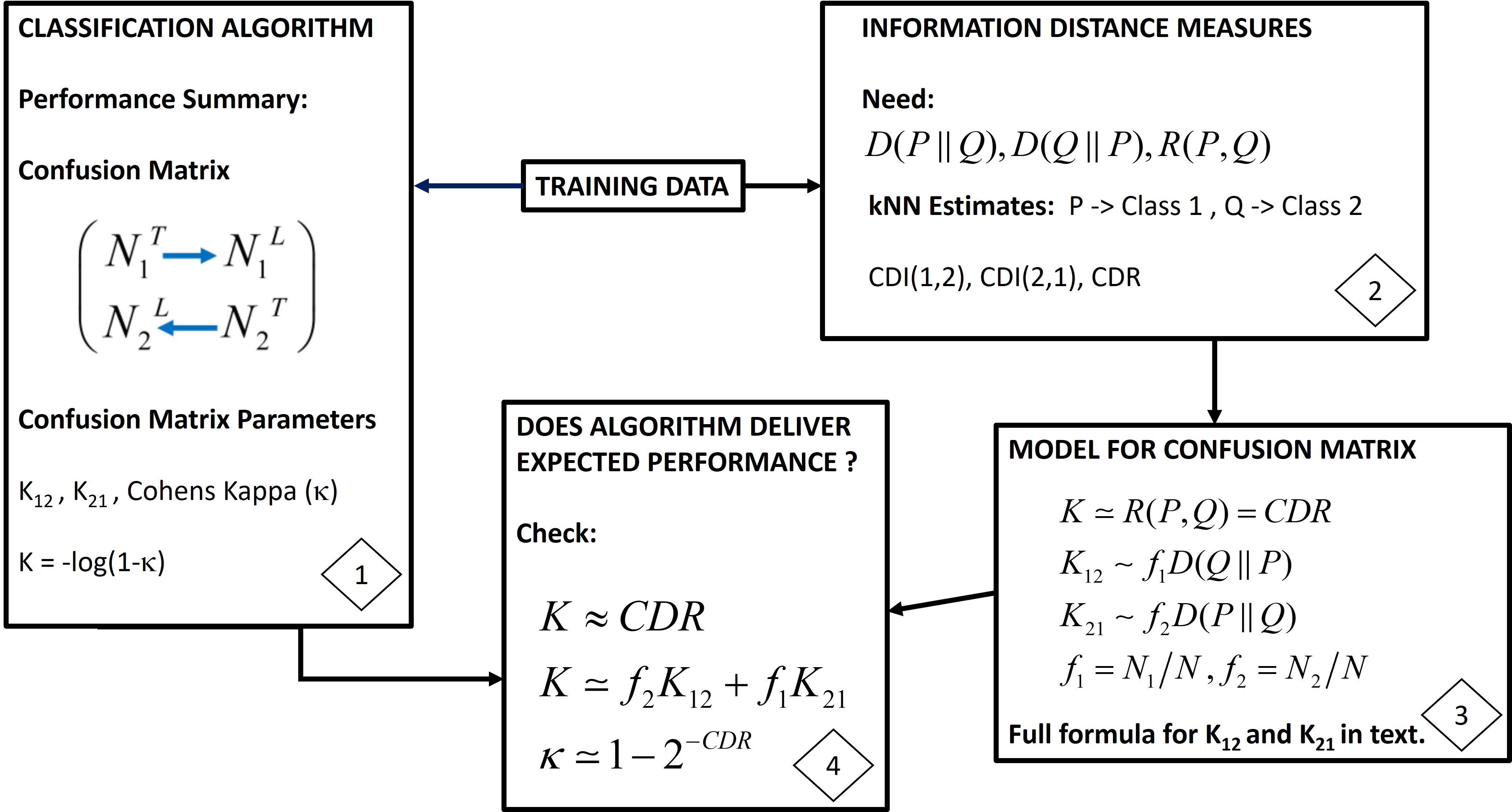}\caption{Methodology to compare performance of a classification algorithm with
expectation from information distance measures. For full details in
the main text; Box 1 see Section 2, Box 2 see Sections 3.1 and 3.3,
Box 3 see Section 3.2, Box 4 see Sections 3.2 and 3.3.}

\label{Figure 2}
\end{figure}

\begin{figure}[p]
\begin{centering}
\includegraphics{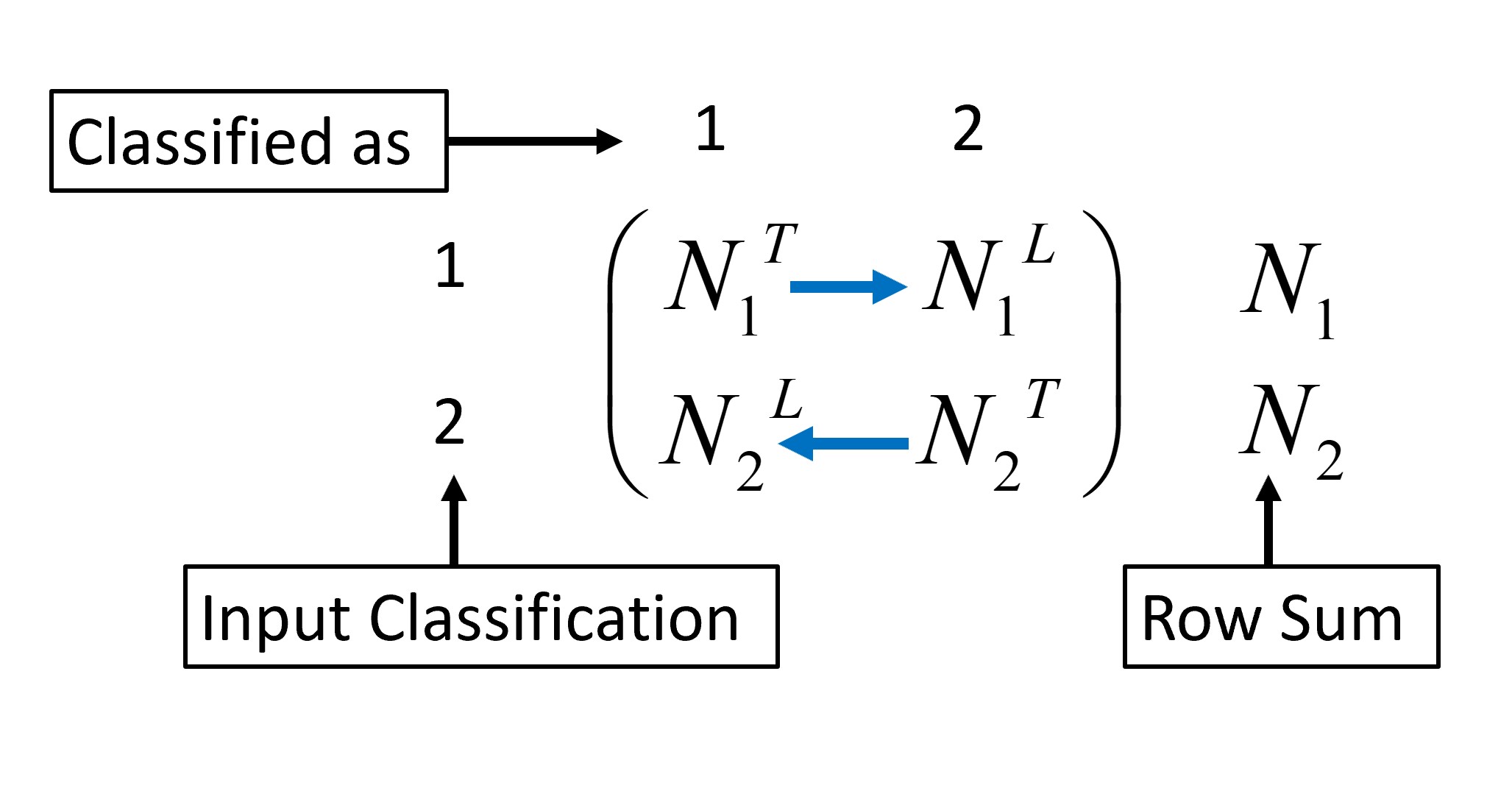}
\par\end{centering}
\caption{Definition of the two class confusion matrix. The arrows indicate
how entries ``leak'' from their true class (T) to the wrong class
(L).}
\label{Figure 3}
\end{figure}

The methodology of this paper is summarised in Fig. \ref{Figure 2}.
The aim was to link the performance of a classification algorithm
on training data with the underlying pdfs. This has been achieved
by estimating key information distance measures from the same training
data, independently of the classification algorithm. These distances
allow the prediction of the best achievable performance which can
be compared to the actual performance. Referring to Fig. \ref{Figure 2},
\begin{itemize}
\item The classification algorithm performance is summarised in terms of
its confusion matrix and Cohen's Kappa. Section 2 showed how the key
parameters, $K$, $K_{12}$ and $K_{21}$ are extracted using Equ.
\ref{eq:2.5}, Equ. \ref{eq:2.6} and Equ. \ref{eq:2.7} respectively. 
\item The relationship between the information measures and the error rates
in Equ. \ref{Eq1:1} and Equ. \ref{Eq1:2} are described in Section
3. The estimates of the information measures for $D(P\parallel Q)$,
$D(Q\parallel P)$ and $R(P,Q)$ are $CDI(1,2)$, $CDI(1,2)$ and
$CDR$ respectively. These are described in Section 3.3. One can also
estimate the information measures by fitting the $K_{12}$ and $K_{21}$
results, extracted from an analysis of the confusion matrix resulting
from the classification algorithm, namely, $D(1,2)$ and $D(2,1)$,
Equs. \ref{eq:3.17-1} and \ref{eq:3.18-1} respectively. 
\item The model for the confusion matrix parameters is described in Section
3.2. The key results are Equ. \ref{eq:3.17-1}, Equ. \ref{eq:3.18-1},
Equ.\ref{eq:3.26} and Equ. \ref{eq:3.27}.
\end{itemize}
Fig. \ref{Figure 2} contains key equations and links to the text.
The next section uses these results to evaluate how well this theory
and model work on simulated and real datasets. 

\section{Application of the theory to data}

Table 1 is a summary of the six datasets, refs. \cite{WattsCrow,BreastC,Bank,CHD},
to which the theory and model described in Section 2 and Section 3
are applied. The key equations are summarised in Fig. \ref{Figure 2}
and Section 3.4. The datasets are of three types. S1 and S2 are Monte
Carlo generated data based on Gaussian and Exponential distributions.
D1, D3 and D4 are datasets available on public websites. D2 has been
used before and is described in Ref. \cite{WattsCrow}. S1, S2, D1
and D3 use continuous variables. D2 has only discrete variables. D4
is a mixed dataset with one discrete variable. These datasets are
described in Section 4.1 and 4.4 and then analysed in Sections 4.2,
4.3, and 4.5. Version 3.9.6 of the WEKA, \cite{Weka}, machine learning
software was used. The classification algorithm used in WEKA is given
in Table 1. The algorithm was chosen on the basis of its performance
with each dataset using standard defaults in WEKA.

\subsection{Monte Carlo simulation data}

S1 and S2 are balanced datasets with $f_{1}=f_{2}=0.5$. They were
generated because one can solve these problems numerically from the
equations defining the pdfs. Each dataset has 16 variables, each of
which uses the same pdf, and these variables are generated independently.
There is no shared mutual information between the variables. The S2
exponential model was used to calculate the information measures that
are shown in Fig. \ref{Figure1B}. The mathematical form is given
in Equ. \ref{eq:4.2}, from which the divergences were calculated
numerically. A Bayes Network classification algorithm was chosen to
process the data. For both datasets the Bayes Network algorithm showed
the WEKA software had correctly identified that all variables were
only dependent on the class variable. This algorithm should perform
optimally for these datasets. If $p_{i}(x)$ and $q_{i}(x)$ are the
pdf's for the $i^{th}$ variable for class 1 and class 2 respectively,
with $i=1$ to $16$, then for the Gaussian dataset S1, the distributions
are,
\begin{equation}
p_{i}(x)=N(0,1),q_{i}(x)=N(1.02,1)
\end{equation}
where $N(\mu,\sigma)$ is a Gaussian distribution with mean, $\mu$
, and standard deviation, $\sigma$. With this choice one can show
that for a single variable, $D(P\parallel Q)=D(Q\parallel P)=0.75$
bits and $R(P,Q)=0.375$ bits. For the exponential distribution the
choice of pdf's are, 
\begin{equation}
p_{i}(x)=\frac{1}{\alpha}exp(-\frac{x}{\alpha}),q_{i}(x)=\frac{1}{\beta}exp(-\frac{x}{\beta})\label{eq:4.2}
\end{equation}
with $\alpha=1$ and $\beta=2.392.$ With this choice one can show
that for a single variable, $D(P\parallel Q)=0.42$ bits, $D(Q\parallel P)=0.75$
bits and $R(P,Q)=0.27$ bits. Due to independence, as the variables
are combined, the total $D(P\parallel Q),D(Q\parallel P)$ and $R(P,Q)$
will be the values stated multiplied by the number of variables used.
This allowed 16 different values of increasing dimensionality to be
processed by both the information distance measure software as well
as the corresponding classification results when using WEKA. Table
2 tabulates which figures show the results from processing the simulated
data; V1 means one variable, V1V2 means the two variables, V1 and
V2, etc. Generating such data uses well known techniques. Ref. \cite{Hist}
provides more detail.

\subsection{Analysis of balanced Monte Carlo data }

\begin{figure}
\includegraphics[scale=0.5]{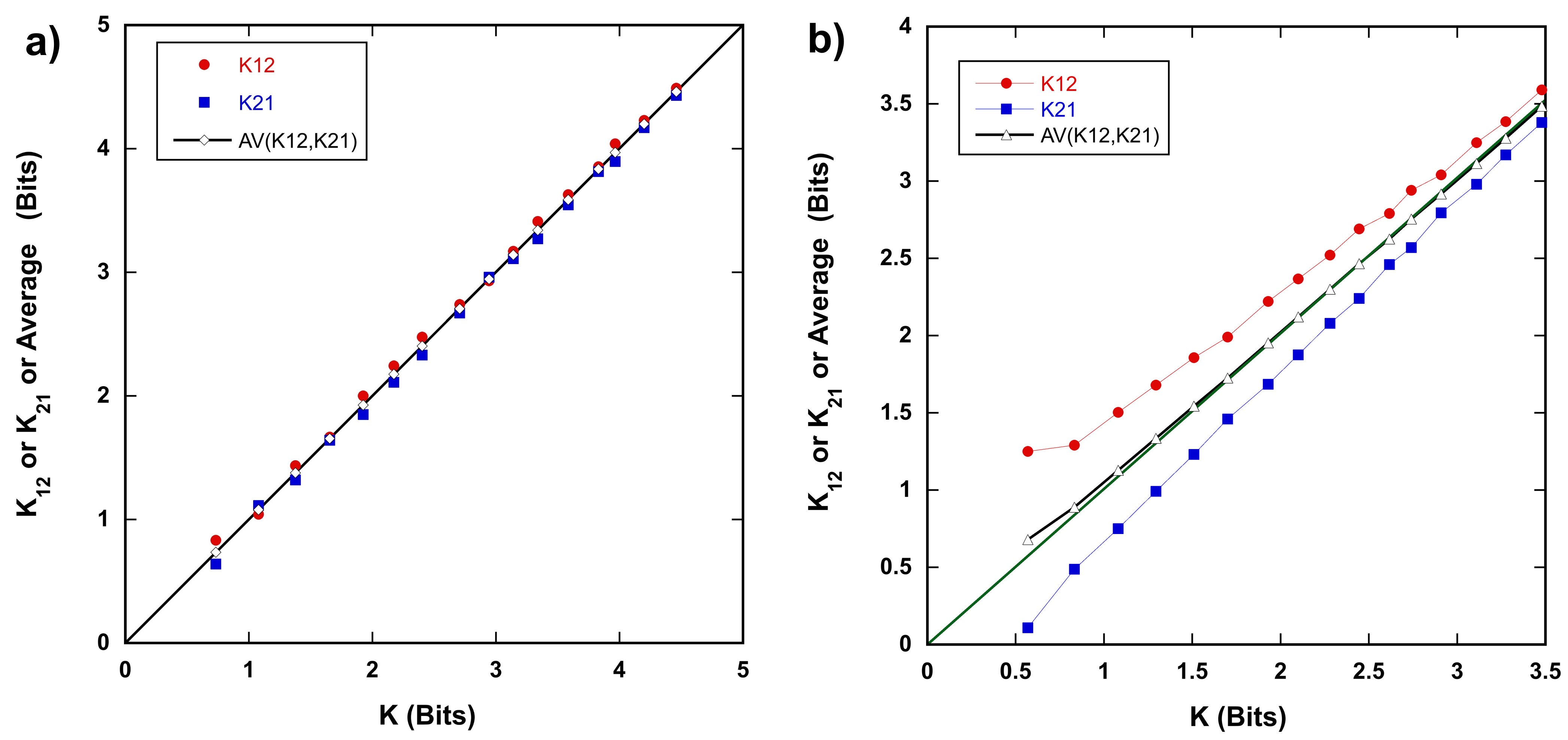}

\caption{a) Relationships between the parameters $K$, $K_{12}$ and $K_{21}$
and average ($\nicefrac{1}{2}(K_{12}+K_{21})$) for the Gaussian Model.
Slope of line is 1.0 for all. b) Repeat for the Exponential Model.
Slope of line for average is 1.0. This is for balanced data, $f_{1}=f_{2}=0.5$.
The models are described in Section 4.1.}

\label{Figure4}
\end{figure}

The Monte Carlo models have $N_{1}=N_{2}=\frac{N}{2}$ , which is
referred to as \emph{balanced} classes. Fig. \ref{Figure4}a) shows
the relationship between $K$, $K_{12}$, and $K_{21}$ for the Gaussion
simulated data described in Section 4.1. The average of $K_{12}$
and $K_{21}$ is also plotted. Since the Gaussian model has a symmetric
Kullback-Leibler Divergence and $f_{1}=f_{2}=0.5$ one finds that
$K=K_{12}=K_{21}=\frac{1}{2}(K_{12}+K_{21})$, as expected. Fig. \ref{Figure4}b)
repeats the exercise for the exponential simulated data described
in Section 4.1. The Kullback-Leibler Divergence is not symmetric and
$K_{12}\neq K_{21}$ because $D(Q\parallel P)\neq D(P\parallel Q).$
As expected, the average, $\frac{1}{2}(K_{12}+K_{21})$ equals $K.$ 

\LyXZeroWidthSpace{}

\begin{figure}

\includegraphics[scale=0.5]{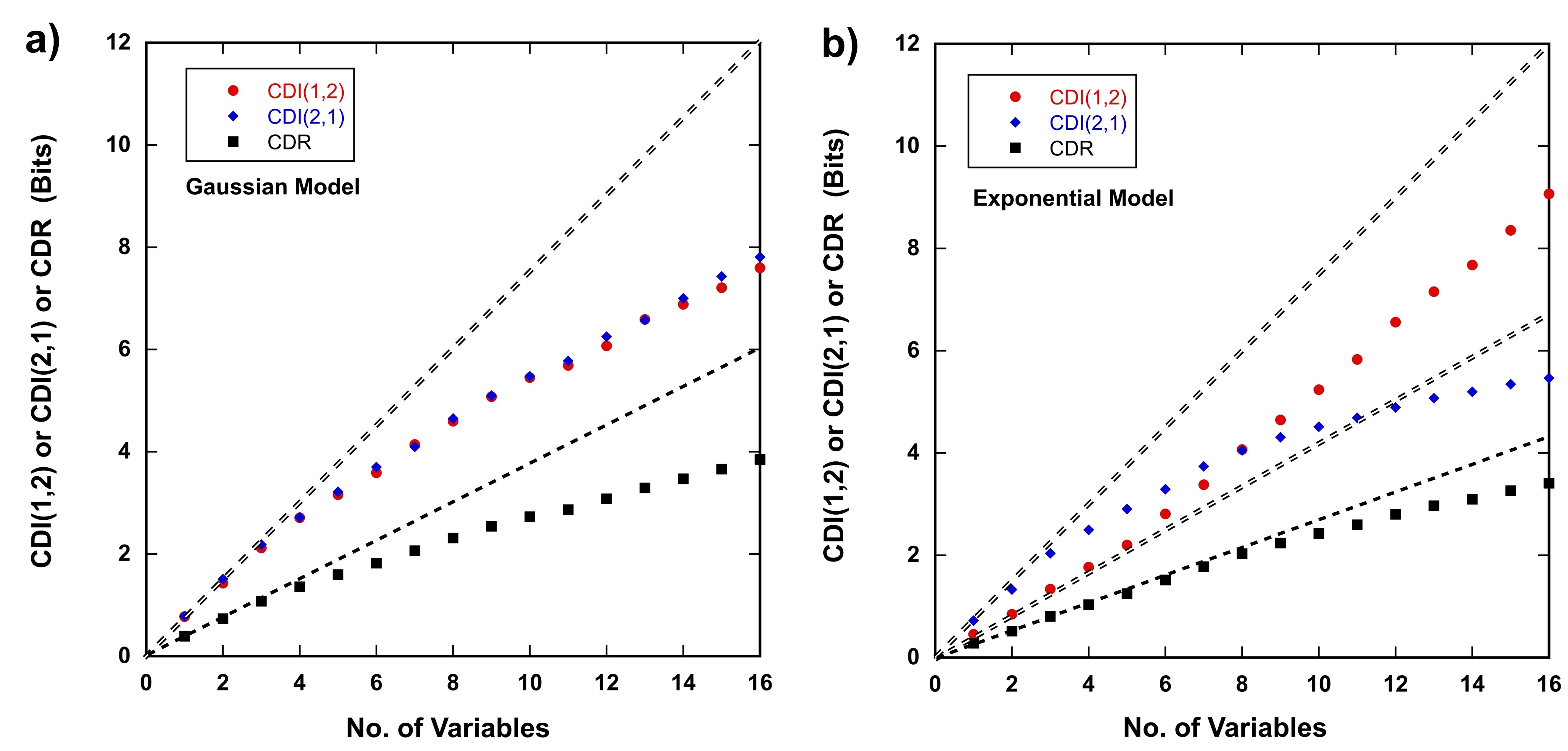}\caption{kNN estimates of the Kullback-Leibler Divergences, $CDI(1,2)$ and
$CDI(2,1)$, and Resistor Average Distance, $CDR$ , for both the
Gaussian and Exponential models. See Sections 3.3 and 4.2 for detail.
Double dashed lines are the theoretical prediction for the Kullback-Leibler
divergence. Single dashed lines are the theoretical prediction for
the Resistor Average Distance.}

\label{Figure5}
\end{figure}

Fig. \ref{Figure5} shows the estimated Kullback-Leibler Divergences,
$CDI(1,2)$, $CDI(2,1)$, and Resistive Average Distance, $CDR$,
for the a) Gaussian model and b) Exponential model. The lines show
the expected theoretical values which are linear in the number of
variables because these they are independent. $CDI$ and $CDR$ use
a double and single line respectively. For the Gaussian model $CDI(1,2)=CDI(2,1)$
since the Kullback-Leibler Divergence is symmetric. Fig. \ref{Figure5}
a) shows that the $CDI$ and $CDR$ are lower than expected for more
than 5 variables. This is due to the unfortunately named ``\emph{curse
of dimensionality}'', ref. \cite{BellmanCurse}. There are two related
aspects to this problem. First, the number of samples required to
estimate an arbitary function to a given accuracy grows exponentially
with dimensionality. Points are most likely to be found close to the
surface and the full phase space is not used. Second, ref. \cite{BeyerkNN},
the distance of a point to its nearest neighbour tends to a constant
as the dimensionality increases. The Kullback-Leibler Divergence depends
on the ratio of two nearest neighbour distances, Equ. \ref{eq:3.16},
so is affected as the number of variables increases. This problem
has been studied by the authors and a correction found. Due to the
extra pages required this is not described here especially as it is
a small perturbation to the main aim of this paper. Moreover, it is
found that the classification algorithms also suffer from the curse
and the uncorrected $CDR$ remains a reliable estimate of the underlying
error rate which is demonstrated by analysis of the datasets described
in this paper. Fig. \ref{Figure5}b) shows the calculation for the
exponential model. As the underlying Kullback-Leibler divergences
are different, the behaviour of $CDI(1,2)$ and $CDI(2,1)$ are quite
different due to the dimensionality. Below 8 variables, $CDI(2,1)>CDI(1,2)$
as expected. However, beyond 8 variables, the relationship switches.
This does not occur to the confusion matrix parameters, $K_{12}$
and $K_{21}$ which remain correctly ordered for all dimensions. The
$CDR$ remains a good estimate of the overall error rate. 

\subsection{Imbalanced Monte Carlo data}

\begin{figure}
\includegraphics[scale=0.5]{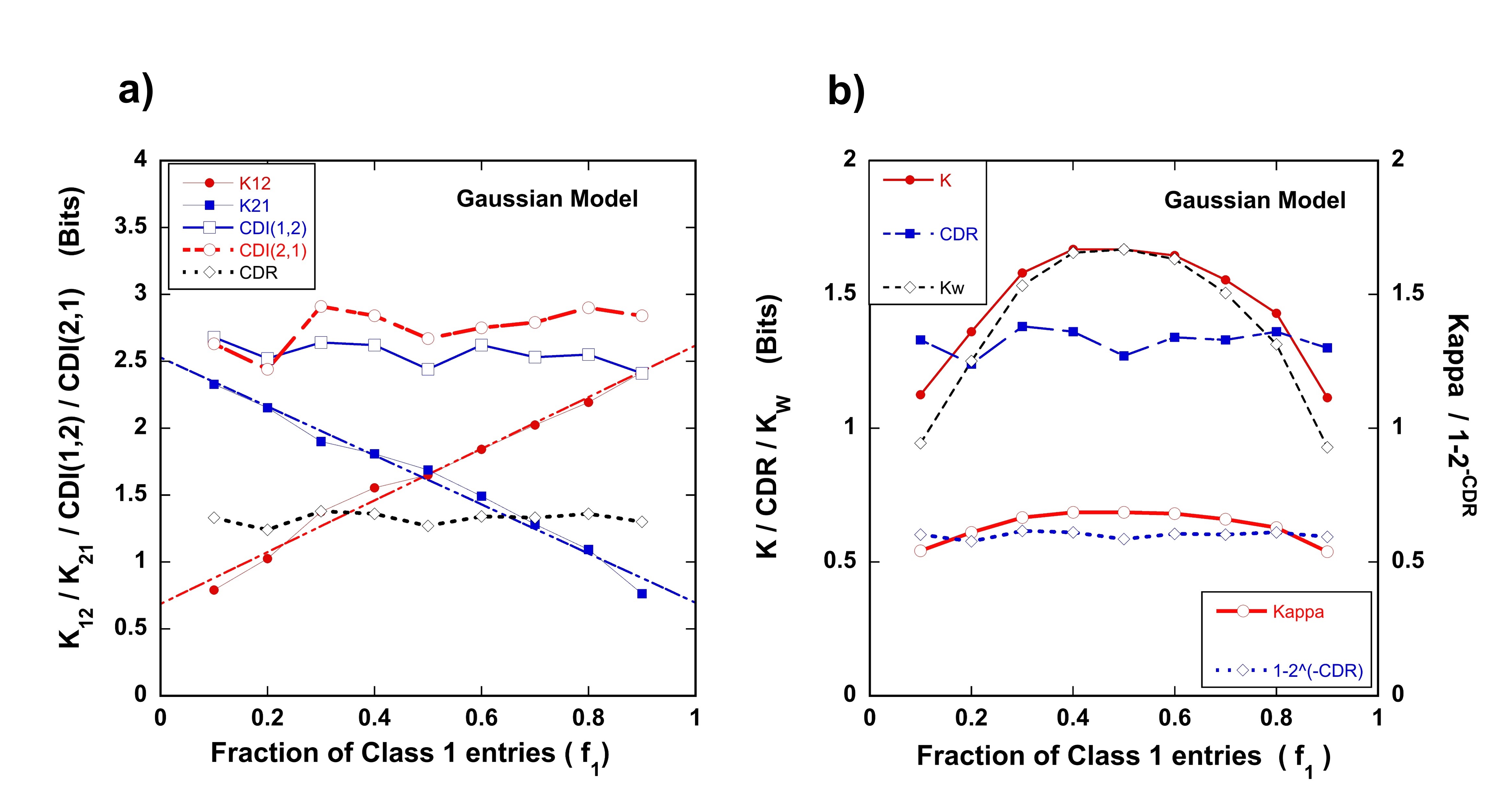}

\caption{a) $K_{12},K_{21},CDI(1,2),CDI(2,1)$ versus the fraction of Class
1 entries for the Gaussian Model. This illustrates how the leakage
rates tend to the Kullback-Leibler Divergence for $D(Q\parallel P)$
and $D(Q\parallel P)$ for $f_{1}=1$ and $f_{1}=0$ respectively.
This is consistent with Equ. \ref{Eq1:1} which is a statement of
the Chernoff-Stein Lemma. b) Left-hand axis, $K$, $CDR$, and the
weighted error rate, $K_{W}$. Right-hand axis, Cohen's Kappa, $\kappa$,
and $1-2^{-CDR}$. Both versus the fraction of Class 1 entries for
the Gaussian Model. The variables on the right-hand axis have a maximum
value of one but the range is set to avoid a clash between the left-hand
and right-hand points. There is reasonable agreement between $K$
and $CDR$. There is better agreement between Cohen's Kappa and $1-2^{-CDR}$.
There is good agreement between $K$ and $K_{W}$ supporting Equ.
\ref{eq:2.9} . }

\label{Figure6}
\end{figure}

\begin{figure}
\includegraphics[scale=0.52]{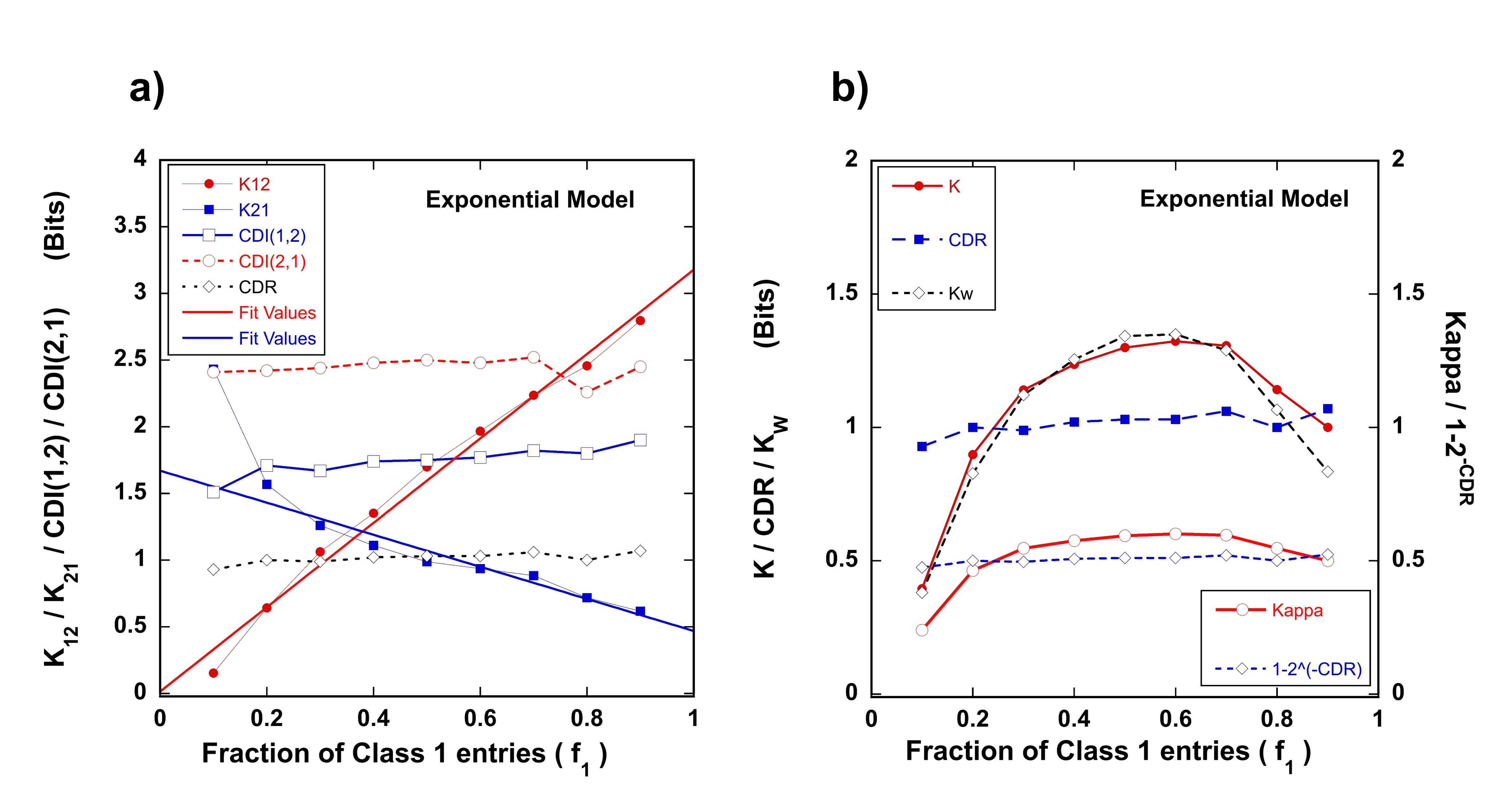}

\caption{Repeat of Fig. \ref{Figure6} for the Exponential Model. a) $K_{12},K_{21},CDI(1,2),CDI(2,1)$
versus the fraction of Class 1 entries. b) Left-hand axis, $K$, $CDR$,
and the weighted error rate, $K_{W}$. Right-hand axis, Cohen's Kappa,
$\kappa$ , and $1-2^{-CDR}$ . Both versus the fraction of Class
1 entries. The variables on the right-hand axis have a maximum value
of one but the range is set to avoid a clash between the left-hand
and right-hand points. There is reasonable agreement between $K$
and $CDR$. There is better agreement between Cohen's Kappa and $1-2^{-CDR}$.
There is good agreement between $K$ and $K_{W}$ supporting Equ.
\ref{eq:2.9} . }

\label{Figure7}
\end{figure}

Real data rarely has balanced classes, a situation called \emph{imbalanced
classes,} i.e. $f_{1}\neq f_{2}$. Moreover, the imbalance in the
training data may be very different to real life. For example, $f_{1}=0.37$
for the Breast Cancer data, but in reality the probability for such
a cancer in the general population is considerably lower\emph{.} The
uncorrected rate for women in the UK between 2016-18 was 166 per 100,000
of the population, ref. \cite{BCStat}. The probability for women
to be diagnosed with this cancer, during their lifetime, is about
1 in 7. When either $f_{1}$ or $f_{2}$ are 0.01 or smaller\emph{,
}the data set is said to be in\emph{ Severe Imbalance}. This is a
well known problem, and care must be taken to train the classification
algorithm for such a situation, see ref. \cite{imbalance}. Cohen's
Kappa is sensitive to imbalance. Equ. \ref{eq:3-1} clearly shows
the $N_{1}$ and $N_{2}$ dependence which is irrelevant when $K_{12}=K_{21}$
or the classes are balanced, $N_{1}=N_{2}$. In Fig. \ref{Figure4}
the classes were balanced for both the Gaussian and Exponential models.
The exercise of obtaining the confusion matrix parameters and information
measures was repeated for different values of $f_{1}$ for both the
Gaussian and Exponential models with four variables only. From Section
4.1, one would expect that $CDI(1,2)=CDI(2,1)=4\times0.75=3.0$ bits
for the Gaussian model. For the Exponential Model one would expect
$CDI(1,2)=0.42\times4=1.68$ bits and $CDI(2,1)=0.75\times4=3.0$
bits. Fig. \ref{Figure6}a) shows the confusion matrix parameters,
$K$, $K_{12}$ and $K_{21}$ and the information distance measures,
$CDI(1,2)$ and $CDI(2,1)$, versus the fraction of Class 1 entries,
$f_{1}$. This figure is symmetric about $f_{1}=0.5$ because $D(P\parallel Q)=D(Q\parallel P)$.
The information distance measures are independent of $f_{1}$ showing
that the correction factor in Equ. \ref{eq:3.16} works well. There
is remarkable resemblance between Fig. \ref{Figure6} and Fig. \ref{Figure1B}
with $f_{1}$ clearly related to the parameter $t.$ This is not a
surprise because $f_{1}$= 0 corresponds to the error rate being maximal
for Class 2 entries, which is the Chernoff-Stein Lemma, Equ. \ref{Eq1:1},
with $K_{21}=D(P\parallel Q)=CDI(1,2)$. The situation is reversed
for $f_{1}=1$ or $f_{2}=0$ with $K_{12}=D(Q\parallel P)=CDI(2,1)$.
$K_{12}$ and $K_{21}$ vary linearily between the two extremes and
are equal at $f_{1}=f_{2}=0.5$. For severely imbalanced classes,
one can apply the Chernoff-Stein Lemma, Equ. \ref{Eq1:1} but not
Equ. \ref{Eq1:2}. An overall error rate only makes sense when one
is trying to identify both classes as well as possible and they must
be reasonably balanced. In this case, this corresponds to $0.2<f_{1}<0.8$.
Fig. \ref{Figure6}b) shows parameters that describe the overall performance
derived from the confusion matrix - $\kappa$ and $K$ - and the information
distance measure, $CDR.$ In the region of reasonable balance, there
is agreement between $K$ and the CDR. Using Equ. \ref{eq:3.27},
the expected value of $\kappa$ can be plotted. There is good agreement
between this and the actual $\kappa$ in the region of reasonable
balance. The weighted error rate, $K_{W}$, Equ. \ref{eq:2.9}, is
also shown in Figs. \ref{Figure6} b) and \ref{Figure7} b). This
agrees remarkably well with the actual $K$. The parametrisations
for $K_{12}$ and $K_{21}$ were then fitted to both Figs. \ref{Figure6}a)
and \ref{Figure7}a). The results are shown in Table 3. Table 4 gives
the information distance measures for the same datasets. There is
good agreement between the various estimates and the expected theoretical
values. This shows that the underlying theory and model is working
well. There is good agreement between the balance points ($K_{12}=K_{21}$)
and that expected from Equ. \ref{eq:3.19} and Equ. \ref{eq:3.24},
confirming again the link between $f_{1}$ and Chernoff's $t$. 

\subsection{Real datasets}

As the simulated data has independent variables the choice of which
to combine and use is simple. In real datasets the variables share
information and thus the choice of which combinations to show is problematic.
There are $2^{p}-1$ combinations for $p$ variables. Only for the
D2 dataset have all been processed. In other cases, most of the single
variables, two variable combinations and selected combinations were
processed. These are detailed in Table 2. The selected variables were
chosen as follows. First, find the single variable with the largest
$CDR.$ Combine this with the next variable which increases $CDR$
the most. Continue this process until the $CDR$ is unchanged by adding
unused variables. One then has a list of $N_{s}$ selected variables
ordered by increasing $CDR$. The number of combinations to check
are $\frac{1}{2}(p-1)(p+2)$, which is considerably less than before.
Moreover, the classification algorithm is only applied $N_{s}$ times.
This paper is not a detailed analysis of the datasets. They are being
used to confirm the methodology. Table 1 contains references to each
data source. Brief information for each dataset is given below.
\begin{enumerate}
\item The breast cancer data, D1, has 30 continuous variables, ref. \cite{BreastC}.
There is a high degree of correlation between the variables and the
number needed to successfully classify the data was found to be 7.
This data has been widely studied and other choices of variables have
been found. Due to the shared information between the variables this
is not surprising. The selection algorithm discussed above gave the
following ordered list of variables; perimeter worst, concave points
mean, radius worst, radius mean, perimeter mean, area mean, texture
worst. The class variable is Malign (M) or Benign (Be).
\item The bankruptcy data, D2, has 6 discrete variables, ref. \cite{Bank}.
All 63 combinations were processed for classification results and
information distance measures. The class variable is Bankrupt (Bk)
or Not Bankrupt (NBk).
\item The Particle Physics data, D3, has 8 continuous variables, ref. \cite{WattsCrow}.
It is the result of a sophisticated Monte Carlo simulation to understand
real physics data. The selection algorithm gave the following ordered
list of variables; Fsig, Sfl, Rxy, Rz, Doca, Cos-Hel, Pchi, Mass.
The final variable was added last because in the real experiment one
would use the classification to identify the particle, and then confirm
its mass. This variable would not be used in the initial classification
algorithm. The class variable is Signal (S) or Background (B).
\item The Coronary Heart Disease data, D4, is the result of a health study
in South Africa. This has 8 continuous and 1 discrete variable, ref.
\cite{CHD}. As will be shown, one cannot achieve a high classification
performance with this data. All single variable, all double variable,
and a mix of 3 variable and 4 variable combinations were processed.
Higher variable combinations actually give worse performance and were
not included. The class variable is No Disease (ND) or Coronary Heart
Disease (CHD).
\end{enumerate}

\subsection{Key results and discussion}

\subsubsection{Full application of the model to one variable in the breast cancer
dataset}

\begin{figure}

\includegraphics[scale=0.5]{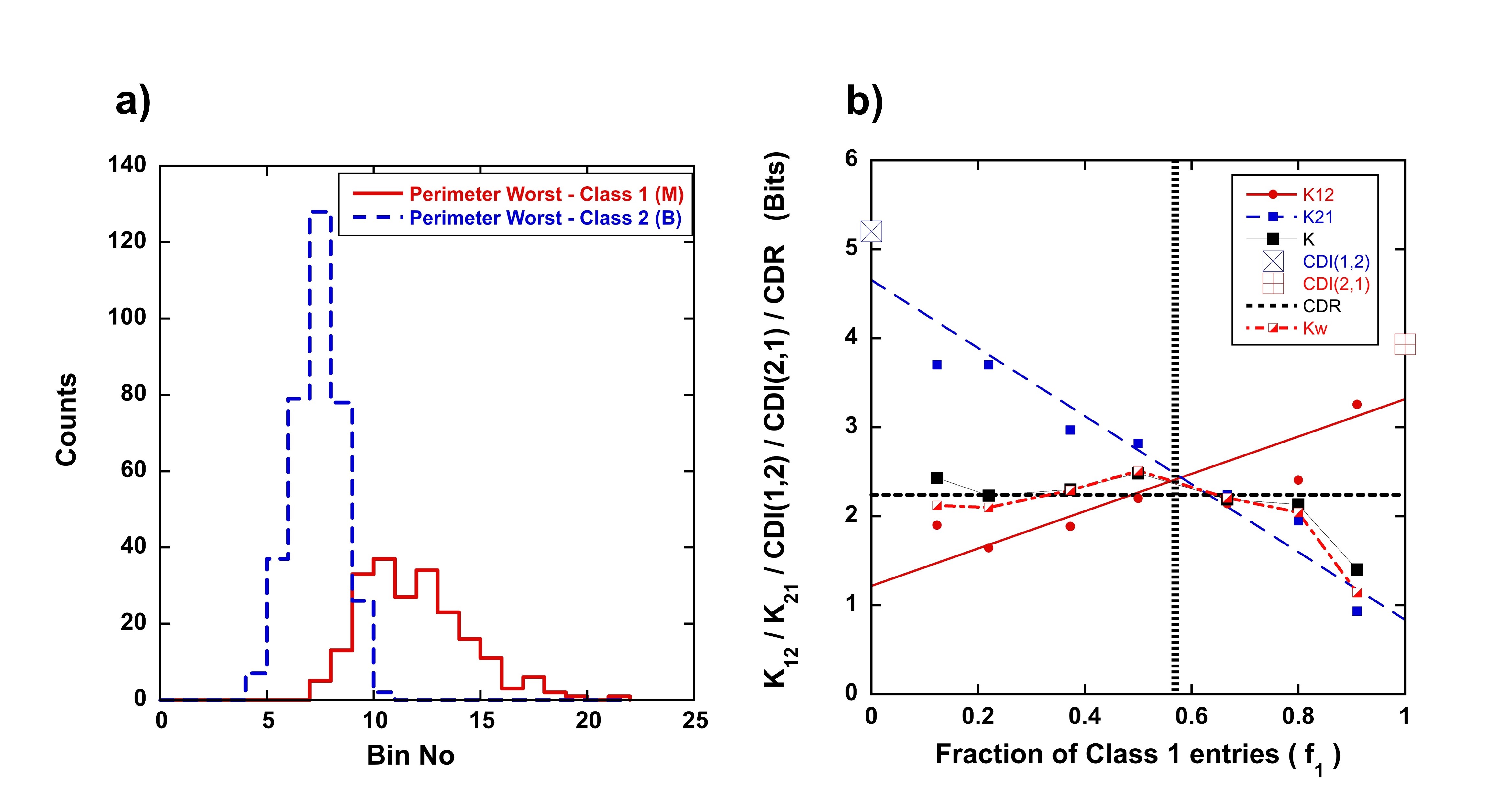}\caption{This figure shows an analysis of the ``Perimeter Worst'' variable
for the Breast Cancer dataset. a) Histogram of the actual data for
this variable with Class 1 (Malignant) in solid and Class 2 (Benign)
in dashed lines. The bin width is 11.8 and binning starts at zero.
The total number of entries is $569$. b) This plot is similar to
7 a) but uses real data. $f_{B}\simeq t_{R}=\frac{CDI(1,2)}{CDI(1,2)+CDI(2,1)}$
is $0.57$ for this variable and is indicated by the vertical dashed
line. It is not an accident that this is the point at which $K_{12}\simeq K_{21}$
. See text for more detail. }

\label{Figure8}
\end{figure}

Fig. \ref{Figure8} shows the behaviour of $K_{12}$ and $K_{21}$
versus $f_{1}$ for the Breast Cancer dataset for just one variable.
This was chosen because it is real data and illustrates that the methodology
works well even though there are only 579 total entries. The most
important variable, ``perimeter worst'' is analyzed as one can display
its pdf using a histogram, shown in Fig. \ref{Figure8}a). The Kullback-Leibler
divergences were estimated using Equ. \ref{eq:3.16} and found to
be $CDI(1,2)$= 5.2 bits and $CDI(2,1)$= 3.43 bits. The distributions
for the two classes have a different mean, which will give a non-zero
Kullback-Leibler divergence, but it is asymmetric because they have
different variances. This dataset has $f_{1}=0.37.$ Other values
of $f_{1}$ were obtained as follows; $f_{1}<0.37$, $N_{2}$ was
kept at its initial value of $357$ and entries from Class 1 were
reduced; for $f_{1}>0.37$, $N_{1}$ was kept at its initial value
of $212$ and entries from Class 2 were reduced. Analysis of these
different values of $f_{1}$ is shown in Fig. \ref{Figure8}b). This
figure shows $K_{12}$, $K_{21}$, $K$ and $CDR$ as a function of
$f_{1}$. The values of $CDI(1,2)$ and $CDI(2,1)$ are placed at
$f_{1}=0$ and $f_{1}=1$ respectively as this corresponds to a situation
in which one is only interested in performance for one class. The
balance point, $f_{R}\simeq t_{R}=0.57$ , is indicated by a vertical
broken line. For Fig. \ref{Figure8}b) from the fit $\Delta_{1}=(1.32\pm0.24)$
bits , $CDI(2,1)=(3.13\pm0.2)$ bits. $\Delta_{2}=(0.84\pm0.27)$
bits , and $CDI(1,2)=(4.66\pm0.2)$ bits. The independent kNN estimate
of $CDI(1,2)=5.2$ bits and $CDI(2,1)=3.43$ bits. These were included
in the fit to the points in Fig. \ref{Figure8} b) although the results
are compatible within the errors when they are excluded. These results
are included in Table 3 and Table 4. The weighted error rate is also
shown in Fig. \ref{Figure8}b) and again shows good agreement with
the actual $K.$ There is excellent consistency between all these
results which confirm the underlying methodology. The same method
can be used for multiple dimensions. This example was chosen because
one can easily show the pdf for one variable.

\subsubsection{Kappa for the full range of $f_{1}$}

\begin{figure}
\includegraphics[scale=0.7]{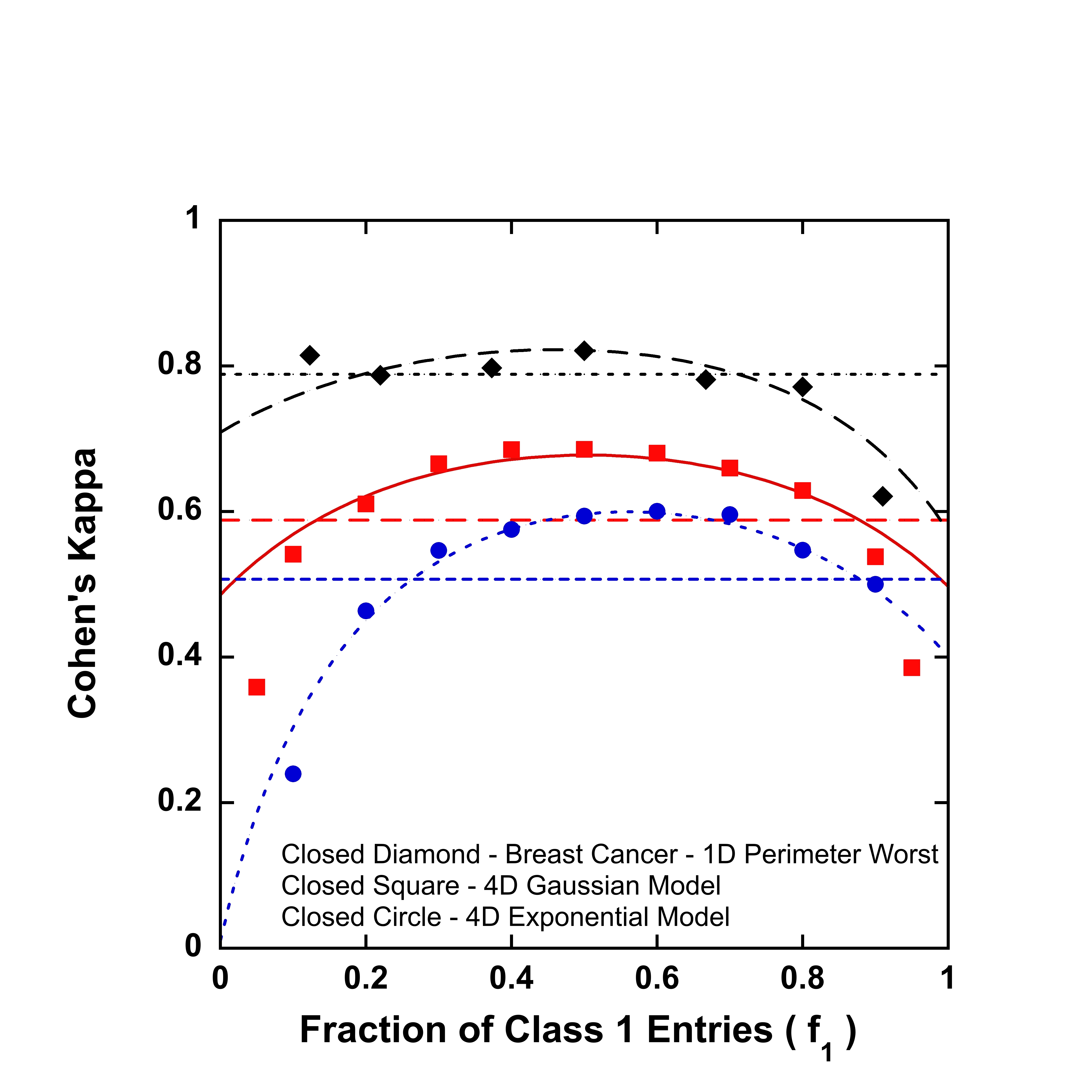}\caption{Cohen's Kappa as a function of the fraction of Class 1 entries for
three datasets. The points are from the classification algorithm.
The curves use the parameters extracted from Figs. \ref{Figure6}a),
\ref{Figure7}a) and \ref{Figure8}b), given in Table 3, and applying
Equ. \ref{eq:3.17-1}, Equ. \ref{eq:3.18-1} and Equ. \ref{eq:4.3}.
$CDR$, the kNN estimate of $R(P,Q),$ is shown by the dashed lines,
and detailed in Table 4. See text for a full discussion.}

\label{Figure9}
\end{figure}

Table 3 provides all the parameters needed to calculate Cohen's kappa
across the full range of $f_{1}$ for two Monte Carlo datasets and
the perimeter worst variable for the Breast Cancer data. One can extrapolate
to find its value at both extremes which would normally not be possible
with small datasets. These parameters plus Equ. \ref{eq:3.17-1} and
\ref{eq:3.18-1} are required. The calculated $K_{12}$ and $K_{21}$
are used to estimate $\kappa$ using Equ. \ref{eq:4.3} , which is
a reformulation of Equ. \ref{eq:3-1} to make the dependence on $f_{1}$
and $f_{2}$ clearer. From Equ. \ref{eq:3-1} one can derive the following
formula which uses units of bits.
\begin{equation}
\kappa=\frac{2(1-f_{1}2^{-K_{21}}-f_{2}2^{-K_{12}})}{2+Z(f_{1}-f_{2})},Z=2^{-K_{12}}-2^{-K_{21}}\label{eq:4.3}
\end{equation}
This formula better illustrates the limits when $f_{1}=f_{2}$ or
$K_{12}=K_{21}$, already discussed in Section 2.2. Using this formula
and the numbers in Table 3, one obtains Fig. \ref{Figure9}. Important
conclusions can be drawn from this figure about the meaning of $\Delta_{1}$
and $\Delta_{2}$, used in Equ. \ref{eq:3.17-1} and Equ. \ref{eq:3.18-1}.
The Gaussian Model has class distributions that are offset with the
same variance. There is an overlap of the classes but not for the
whole range of either class. $\kappa$ is symmetric about $f_{1}=0.5$
and $\Delta_{1}=\Delta_{2}\neq0$ because they do not fully overlap.
$\kappa$ is non-zero at the extremes, and there is some evidence
that the classification algorithm is not working correctly at the
extremes of $f_{1}$. The perimeter worst class distributions are
offset and have different variances. There is an overlap of classes
but not for the whole range. $\kappa$ is not symmetric about $f_{1}=0.5$
. Moreover, $\Delta_{1}\neq\Delta_{2}$ and are non-zero because they
do not fully overlap and consequently, $\kappa$ is non-zero at the
extremes. The Exponential Model has class distributions that overlap.
Class 1 is completely overlapped by Class 2. However, Class 2 has
some part of its range in which the overlap is very unlikely. This
is the reason why $\Delta_{1}=0$ and $\Delta_{2}>0$. Consequently
as $f_{1}\rightarrow0$ , $K_{12}\rightarrow0$ and $\kappa\rightarrow0$.
When the number of Class 1 entries is very low then if they occur
they will be classified as Class 2. A different outcome happens at
$f_{1}$=1 because $\Delta_{2}>0$ then $K_{21}>0$ and $\kappa$
is non-zero. When the number of Class 2 entries is very low then if
they occur the algorithm can still classify some entries correctly
because they are more likely to be outside the range of Class 1 entries.
This shows the importance that pdf's should not fully overlap if the
algorithm is to work at low numbers of entries. This example illustrates
that $\Delta_{1}$ and $\Delta_{2}$ are a measure of whether a class
is fully separated from other classes; $\Delta=0$ means they fully
overlap. 

\subsubsection{Overall average error, $K$, and the Resistor Average Distance for
real data}

\begin{figure}
\includegraphics[scale=0.8]{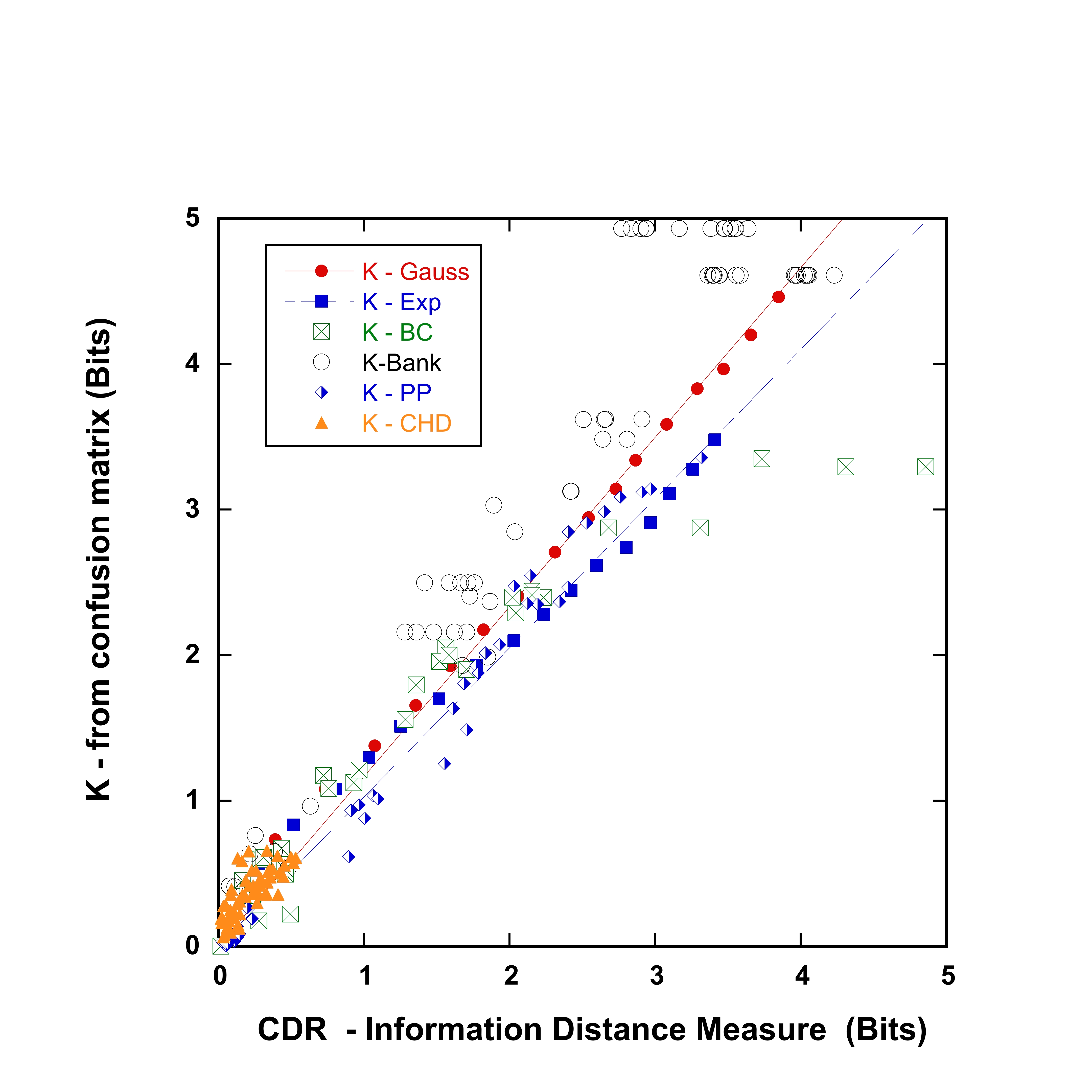}

\caption{Average error rate, $K$, from the confusion matrix Cohen\textquoteright s
Kappa, versus $CDR$ , which is the kNN estimate of the Resistor Average
Distance, $R(P,Q)$. This compares the classification algorithm performance
versus the expected performance from an information theoretic distance.
Gaussian (Closed Circle), Exponential (Closed square), Breast Cancer
data (Crossed Open square), Bank data (Open circle), Particle Physics
data (Half-filled diamond), Coronary Heart Disease data (Closed triangle).
CHD data all below 0.5 bits. Line fits to the Gauss and Exponential
data give slopes of 1.165 and 1.025. }

\label{Figure10}
\end{figure}

Fig. \ref{Figure10} is a key result of this paper. The overall error
rate, described by $K$, which is derived from Cohen's Kappa, $\kappa$,
which is obtained from the confusion matrix describing the result
of the machine learning classification algorithm, is plotted versus
the $CDR$ which is an estimate of the Resistor Average Distance $R(P,Q)$,
and also $D_{\nicefrac{1}{2}}(P\parallel Q)$. Both simulated data
and the real datasets are on this plot. Despite the very different
nature of the datasets, there is a good linear correlation between
these parameters. This figure plus the background theory gives confidence
that Equ. \ref{eq:3.17} applies in general which is also confirmed
by the derivation of Equ. \ref{eq:3.26}.

\subsubsection{Cohen's Kappa versus the Resistor Average Distance for real data}

\begin{figure}
\includegraphics[scale=0.7]{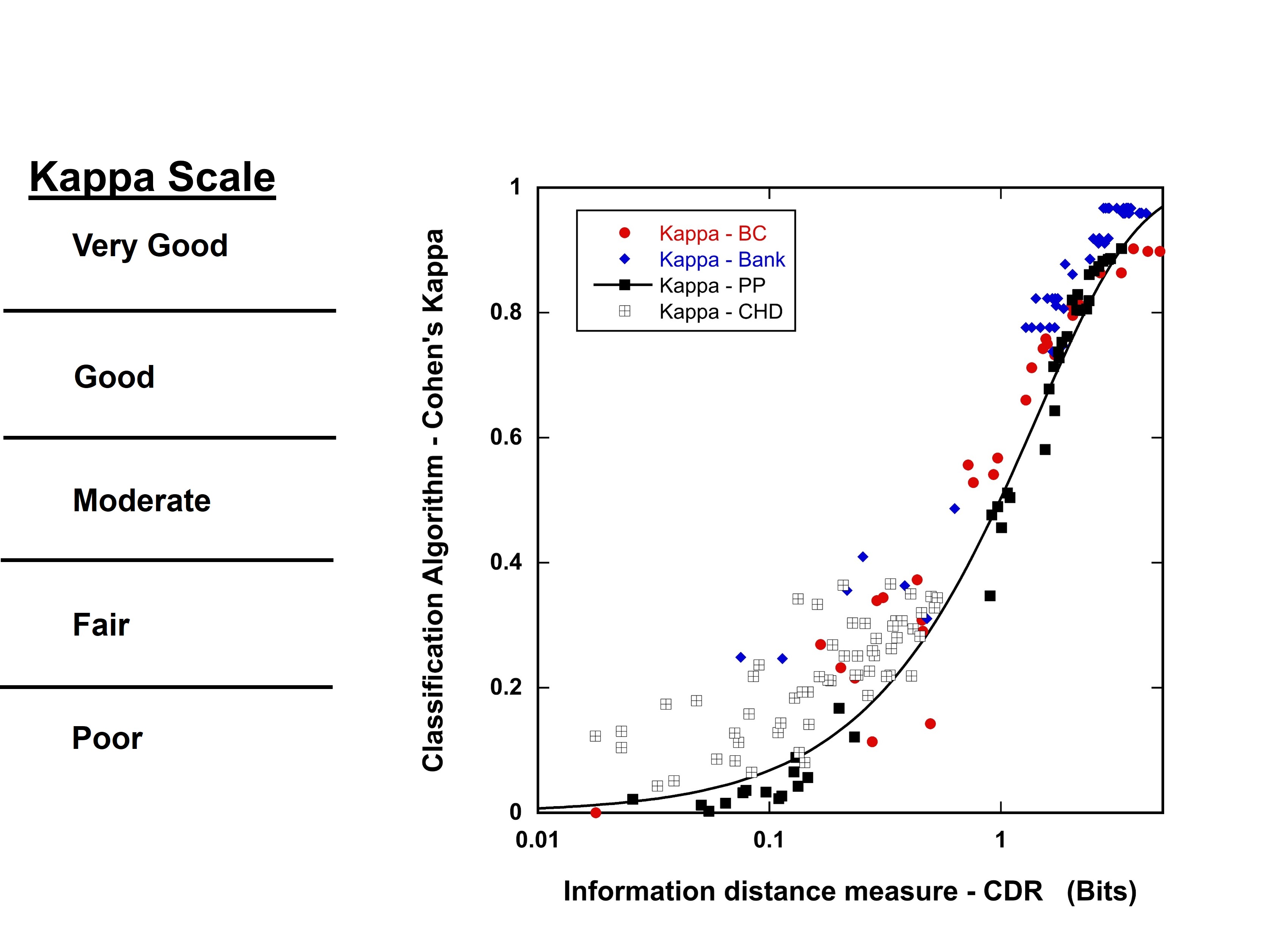}

\caption{This figure combines the result of the classification algorithm performance
using Cohen's Kappa versus the independently calculated, $CDR$, which
is an estimate of the Resistor Average Distance, $R(P,Q)$, which
is an information distance measure. The curve shows the relation,
$\kappa=1-2^{-CDR}$, Equ. \ref{eq:3.27}. The points lie on this
curve which indicates that the classification algorithm is performing
as well as can be expected. The Kappa scale on the left hand side
is from ref. \cite{KappaScale}. The datasets are D1 to D4 detailed
in Table 1.}

\label{Figure11}
\end{figure}

Fig. \ref{Figure11} is the most important result. The goal of the
work was to relate the performance of the classification algorithm
with expectation from the underlying pdf's, using the same training
data. This figure shows this goal has been achieved. Cohen's Kappa,
which is a performance measure of the classification algorithm, agrees
with the $CDR$ information distance measure. The curve in the figure
is Equ. \ref{eq:3.27}. The real datasets have been plotted in this
figure and show excellent agreement with Equ. \ref{eq:3.27}. A wide
range of different numbers of variables with differing kappa were
chosen to show that each dataset agreed across a wide range of $CDR.$
The same classification algorithm was used for any mix of variables.
A kappa scale taken from ref. \cite{KappaScale} shows that all datasets
achieve very good classification, except for the coronary heart disease
data. The inability to classify the heart disease data more effectively
is due to a lack of discrimination in the variables required to achieve
a high $CDR$ and thus a significant kappa. This figure shows there
is no need to search for a new algorithm that might give an improved
result. However, there are links between the variables and coronary
heart disease that give non-zero $CDR$. This data is useful for risk
analysis, which historically was how it was analyzed, which in turn
led to a public health programme that improved outcomes for this disease
in South Africa, ref. \cite{risk}.

\section{Conclusions}
\begin{itemize}
\item To our knowledge, this is the first time that the performance of machine
learning classification algorithms has been compared with an information
distance measure, estimated independently from the underlying probability
density functions of the two classes using the same training data.
The information distance measure provides a best case performance
estimate so one can check that the machine learning algorithm is optimal.
\item The methodology applies to discrete, continuous or mixed data. Performance
for imbalanced classes has been understood and related to the underlying
information distances. The methodology works for any number of variables
although more work is required to understand both the information
distance measures and classification algorithm at high dimensions.
\item The methodology allows one to understand how well an algorithm should
perform when the class imbalance is severe, which is the case for
many real world problems.
\item No algorithm, no matter how clever, can better the performance limit
set by the information distance measures. For example, the coronary
heart disease (CHD) data is only able to deliver \textquotedblleft Fair\textquotedblright{}
performance. This type of data is useful for risk analysis but not
prediction. 
\item The methodology can be applied to multi-class data by taking the classes
in pairs. In principle, the confusion matrix parametrization could
be extended to a multi-class problem.
\end{itemize}

\section*{Acknowledgements}

LC thanks UKRI (STFC) and The University of Manchester for research
student funding. SW thanks the Leverhulme Trust for their support
with an Emeritus Fellowship.

\section*{FIGURES}




\clearpage{}

\section*{Tables}

\begin{wraptable}{l}{0.5\columnwidth}%
\begin{raggedright}
\caption{Summary of Datasets}
\par\end{raggedright}
\begin{tabular}{|l|l|l|l|l|l|l|l|}
\hline 
\textbf{Data} & \textbf{Label} & \textbf{$N_{1}$} & \textbf{$N_{2}$} & \textbf{$N$} & \textbf{$f_{1}$} & \textbf{Continuous /} & \textbf{WEKA Algorithm \cite{Weka}}\tabularnewline
 &  & Class 1 & Class 2 &  &  & \textbf{Discrete} & \tabularnewline
\hline 
\hline 
16 Dim. Gauss & S1 & 8192 & 8192 & 16384 & 0.5 & 16/0 & Bayes Network\tabularnewline
\hline 
16 Dim. Exp. & S2 & 8192 & 8192 & 16384 & 0.5 & 16/0 & Bayes Network\tabularnewline
\hline 
Breast Cancer \cite{BreastC} & D1 & 212 (M) & 357 (Be) & 569 & 0.37 & 30/0 & Simple Logistic Regression\tabularnewline
\hline 
Bankruptcy \cite{Bank} & D2 & 107 (Bk) & 143 (NBk) & 250 & 0.43 & 0/6 & J48 Decision Tree\tabularnewline
\hline 
Particle \cite{WattsCrow} & D3 & 3736 (B) & 1264 (S) & 5000 & 0.75 & 8/0 & Random Forest\tabularnewline
\hline 
Heart Disease \cite{CHD} & D4 & 302 (ND) & 160 (CHD) & 462 & 0.65 & 8/1 & Logistic Regression\tabularnewline
\hline 
\end{tabular}

\end{wraptable}%
\begin{wraptable}{l}{0.5\columnwidth}%
\begin{raggedright}
\caption{Combination of variables used in each figure}
\par\end{raggedright}
\begin{tabular}{|c|c|c|c|}
\hline 
\textbf{Figure} & \textbf{Data} & \textbf{Combinations of variables} & \textbf{Combinations}\tabularnewline
\hline 
\hline 
4a & S1 & V1,V1V2,V1V2V3,.....,V1V2....V15V16 & 16\tabularnewline
\hline 
4b & S2 & V1,V1V2,V1V2V3,.....,V1V2....V15V16 & 16\tabularnewline
\hline 
5a & S1 & As 4a & 16\tabularnewline
\hline 
5a & S2 & As 4b & 16\tabularnewline
\hline 
6 & S1 & 4D - V1V2V3V4 & 1\tabularnewline
\hline 
7 & S2 & 4D - V1V2V3V4 & 1\tabularnewline
\hline 
8 & D1 & 1D - Perimeter Worst & 1\tabularnewline
\hline 
9 & D1,S1,S2 & In order - Perimeter Worst, 4D, 4D & 3\tabularnewline
\hline 
10 & S1 & As 4a & 16\tabularnewline
\hline 
10 & S2 & As 4b & 16\tabularnewline
\hline 
10 & D1 & 26 single variables + Selected; V1V2,V1V2V3,....,V1V2...V7 & 32\tabularnewline
\hline 
10 & D2 & All combinations of 6 variables & 63\tabularnewline
\hline 
10 & D3 & 8 single, 25 pairs + Selected; V1V2V3,.......,V1V2V3....V8 & 42\tabularnewline
\hline 
10 & D4 & 9 single, 36 pairs, 16 x 3 variable, 3 x 4 variable & 64\tabularnewline
\hline 
11 & D1, D2, D3, D4 & As 10 & 201\tabularnewline
\hline 
\end{tabular} \end{wraptable}%
\begin{wraptable}{l}{0.5\columnwidth}%
\begin{raggedright}
\caption{Confusion Matrix parameters - 4.3 and 4.5.1}
\par\end{raggedright}
\begin{tabular}{|l|c|c|c|c|}
\hline 
\textbf{\noun{Data}} & \multicolumn{1}{c}{\textbf{\noun{$K_{12}$ Parameters}}} &  & \multicolumn{1}{c}{\textbf{\noun{$K_{21}$ Parameters}}} & Figs. 6a,7a, 8b\tabularnewline
\hline 
Units - Bits & $\Delta_{1}$ & $D(2,1)$ & $\Delta_{2}$ & $D(1,2)$\tabularnewline
\hline 
\hline 
4D Gauss Model & $0.68\pm0.05$ & $2.62\pm0.05$ & $0.70\pm0.05$ & $2.53\pm0.05$\tabularnewline
\hline 
4D Exponential Model & $0.01\pm0.07$ & $3.18\pm0.07$ & $0.47\pm0.06$ & $1.67\pm0.07$\tabularnewline
\hline 
\multirow{1}{*}{Breast Cancer - PW} & $1.22\pm0.32$ & $3.31\pm0.26$ & $0.84\pm0.27$ & $4.66\pm0.23$\tabularnewline
\hline 
\end{tabular}\end{wraptable}%
\begin{wraptable}{l}{0.5\columnwidth}%
\begin{raggedright}
\caption{kNN parameters - 4.3 and 4.5.1}
\par\end{raggedright}
\noindent \raggedright{}%
\begin{tabular}{|c|c|c|c|}
\hline 
\textbf{Data} & \multicolumn{1}{c}{\textbf{\noun{kNN Parameters}}} & \multicolumn{1}{c}{} & Figs. 6a, 7a, 8b\tabularnewline
\hline 
Units - Bits & $CDI(1,2)$ & $CDI(2,1)$ & $CDR$\tabularnewline
\hline 
\hline 
4D Gauss Model & $2.46$ & $2.66$ & $1.28$\tabularnewline
\hline 
4D Exponential Model & $1.74$ & $2.5$ & $1.02$\tabularnewline
\hline 
Breast Cancer - PW & $5.2$ & $3.93$ & $2.24$\tabularnewline
\hline 
\end{tabular}\end{wraptable}%

\end{document}